%% file: main.tex
\documentclass[10pt]{article}
\usepackage[accepted]{tmlr}

\input{math_commands.tex}

\usepackage{url}
\usepackage[acronym]{glossaries}
\usepackage{subcaption}
\usepackage{paralist}
\usepackage{wrapfig}
\usepackage{amsmath}
\usepackage{amsfonts} 
\usepackage{xfrac}
\usepackage{algorithm}
\usepackage{algpseudocode}
\usepackage{graphicx}
\usepackage[
    colorlinks,
    linkcolor={red!50!black},
    citecolor={blue!50!black},
    urlcolor={blue!80!black}]{hyperref}
\usepackage[utf8]{inputenc}
\usepackage[T1]{fontenc}
\usepackage{booktabs}    
\usepackage{nicefrac}
\usepackage{microtype}      
\usepackage{xcolor}     
\usepackage{float}
\usepackage{cleveref}
\usepackage{tabularx}
\Crefname{section}{\S}{\S\S}
\usepackage[inline]{enumitem}
\glsdisablehyper
\usepackage{multirow}

\newacronym{fips}{FiPS}{\texorpdfstring{\textbf{Fi}ne-grained \textbf{P}arameter \textbf{S}haring}{Fine-grained Parameter Sharing}}

\title{Learning Fine-grained Parameter Sharing\\ via Sparse Tensor Decomposition}

\newcommand{\ucalgary}{Department of Electrical and Software Engineering\\
Schulich School of Engineering, University of Calgary\\
Calgary, AB, Canada}
\newcommand{\tum}{Department of Computer Science\\
Technical University of Munich\\
Bavaria, Germany}

\author{%
  \name Cem Üyük\thanks{Correspondence to: Cem Üyük \href{mailto:cem.ueyuek@tum.de}{<cem.ueyuek@tum.de>}, Utku Evci \href{mailto:evcu@google.com}{<evcu@google.com>}.} \email cem.ueyuek@tum.de \\
  \addr \tum
  \AND
  \name Mike Lasby \email mklasby@ucalgary.ca \\
  \addr \ucalgary
  \AND
  \name Mohamed Yassin \email mohamed.yassin@ucalgary.ca \\
  \addr \ucalgary
  \AND
  \name Utku Evci\thanks{Equal advising.}\hspace{0.4em}\footnotemark[1] \email evcu@google.com \\
  \addr Google DeepMind, Canada
  \AND
  \name Yani Ioannou\footnotemark[2] \email yani.ioannou@ucalgary.ca \\
  \addr \ucalgary
}

\def\month{06}
\def\year{2026}
\def\openreview{\url{https://openreview.net/forum?id=vbS7Z8Zswe}}
\begin{document}\maketitle

\begin{abstract}
Large neural networks achieve state-of-the-art performance on many tasks, yet their sheer size hinders deployment on resource-constrained devices. Among existing compression approaches, cross-layer parameter sharing remains relatively unexplored for transformer models. In this paper, we introduce Fine-grained Parameter Sharing (FiPS), a unified framework for compressing transformer Multi-Layer Perceptrons (MLPs) that combines cross-block parameter sharing, low-rank factorization, and sparsity in a single optimization. FiPS concatenates MLP weight matrices across a group of transformer blocks and factorizes them into a shared basis and sparse, layer-specific projection matrices. Both factors are initialized via singular value decomposition (SVD) and jointly optimized by block-wise reconstruction error minimization. FiPS compresses Vision Transformers (ViTs) by up to 33\% with less than 1\% top-1 accuracy loss on ImageNet-1k, and by up to 57\% when combined with fine-tuning. It also compresses Large Language Models (LLMs) by up to 20\% while outperforming existing SVD-based methods in perplexity and downstream benchmarks at matched compression. Combined with Quantization-Aware Training (QAT), 3-bit FiPS on \textsc{Gemma-2-2B} achieves lower perplexity than 2-bit QAT alone while matching the same 8$\times$ compression. These results establish fine-grained parameter sharing as a practical and effective approach for transformer MLP compression.
\end{abstract}

\begin{figure}[!t]
    \centering
    \includegraphics[width=0.6\textwidth]{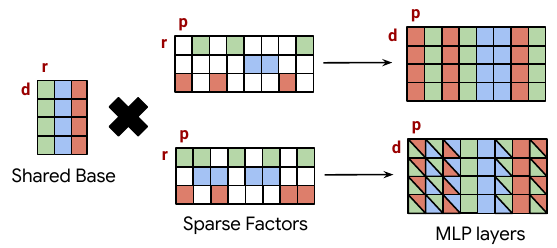}
    \caption{\gls{fips}.}
    \label{fig:sparseonehot}
\end{figure}
\glsresetall

\section{Introduction}
\label{sec:introduction}

Over the past decade, large neural networks have delivered impressive performance through scaling of datasets and model sizes. However, this trend has introduced substantial computational, memory, and storage burdens, highlighting the need for efficient model compression to reduce overhead and enable deployment on resource-constrained devices such as mobile phones and embedded systems. In response, researchers have explored various strategies, including tensor decomposition, quantization, distillation, sparsity, adaptive computing methods, and \emph{parameter sharing}~\citep{ModelCompressionSurveyCheng}. While most of these techniques are well-studied and widely adopted, parameter sharing in neural networks remains underexplored and has not yet been leveraged successfully to compress Transformer models, despite its potential for significant parameter-count reduction. Transformer architectures are a particularly natural target for parameter sharing because they are composed of repeated homogeneous blocks whose Multi-Layer Perceptron (MLP) modules have identically shaped fully connected (FC) layers across depth, enabling direct concatenation and joint factorization of weights across blocks.

Sharing parameters across layers reduces memory usage and improves cache efficiency. Prior works have investigated reusing entire Transformer blocks~\citep{ALBERT, LessonsParamShareTransformers, UnderstsandParamSharingTransformers}, yielding more efficient models. Although directly sharing unmodified weights is promising, we hypothesize that a more fine-grained approach could yield superior compression. 

We therefore focus on sharing neurons across layers by introducing a shared basis, with each neuron expressed as a linear combination of basis vectors via a projection matrix. We, then, find that enforcing sparsity in the projection matrix is essential for the effectiveness of this approach. This insight leads to our novel parameter sharing algorithm, \gls{fips}, which, as we demonstrate, effectively compresses large Vision Transformers (ViTs) and Large Language Models (LLMs)\footnote{All model links are available in~\Cref{app:models}.}. Our contributions include:

\begin{itemize}
    \item \textbf{Systematic Analysis of Cross-Block MLP Sharing.} We systematically explore strategies for sharing bases and neurons across transformer MLP modules, examining sharing granularity, concatenation schemes, and sparsity patterns. We identify when neuron sharing is most effective and how global versus local and structured versus unstructured sparsity interact with cross-block sharing.
    \item \textbf{\gls{fips} Algorithm\footnote{Code: \url{https://github.com/cemuyuk/FiPS}.}.} A shared basis and sparse layer-specific projection matrices are initialized via SVD and jointly optimized by block-wise reconstruction error minimization, followed by optional end-to-end fine-tuning (FT).
    \item \textbf{State-of-the-Art ViT \& LLM Compression.} \gls{fips} delivers substantial compression with minimal performance loss, surpassing recent baselines. It compresses \textsc{DeiT-B} and \textsc{Swin-L} by up to 33\% with $<$1\% top-1 accuracy drop across five vision benchmarks (up to 57\% with fine-tuning), and compresses \textsc{Llama-7B} and \textsc{Llama-3.1-8B} by up to 20\% while outperforming existing SVD-based methods on 10 NLP benchmarks.
    \item \textbf{Quantization-Aware Training (QAT).} We demonstrate that 3-bit QAT combined with \gls{fips} compresses \textsc{Gemma-2-2B} effectively, achieving the same 8$\times$ compression as 2-bit quantization but with markedly better language modeling. This confirms that \gls{fips} is orthogonal to QAT.
\end{itemize}

The remainder of this paper is organized as follows: Section~\ref{sec:ParamSharing} formalizes the integration of low-rank factorization and sparsity for parameter sharing; Section~\ref{sec:method} details the \gls{fips} algorithm; Section~\ref{sec:MainResults} presents our empirical findings; and Section~\ref{Sec:Ablation} provides ablation studies on key design choices. Extensive supplementary material is provided in the Appendix, including additional experiments, algorithmic details, and hyperparameter sweeps.

\section{Parameter Sharing Through Sparse Tensor Decomposition}\label{sec:ParamSharing}
Consider a weight matrix $\mathbf{W} \in \mathbb{R}^{d \times p}$, where $p > d$, that projects feature vectors from a $d$-dimensional space to a $p$-dimensional space with neurons represented by the columns of $\mathbf{W}$. Our objective is to share weights among a subset of these $p$ neurons, reducing the number of unique neurons to $r < p$. In other words, only $r$ columns of $\mathbf{W}$ will contain unique values. These $r$ unique neurons are represented using a shared basis $\mathbf{U} \in \mathbb{R}^{d \times r}$, which is orthogonal at initialization (via SVD) when $r \le d$. The original matrix $\mathbf{W}$ is then reconstructed by mapping each of its $p$ columns to an $r$-dimensional one-hot vector via a projection matrix $\mathbf{V} \in \mathbb{R}^{r \times p}$, i.e., $\mathbf{W} = \mathbf{U}\mathbf{V}$. This ``one-hot'' approach is illustrated in the upper part of~\Cref{fig:sparseonehot}. However, limiting the number of unique neurons to $r$ constrains the representational capacity of $\mathbf{W}$. To address this limitation, we can increase the number of non-zero elements in $\mathbf{V}$, effectively creating linear combinations of basis neurons and generating a significantly larger set of unique neuron representations, as shown in the lower part of \Cref{fig:sparseonehot}. 

This approach readily extends from sharing neurons within a single weight matrix $\mathbf{W}$ to multiple weight matrices $\mathcal{W} = \{\mathbf{W}_1, \ldots, \mathbf{W}_N\}$ via concatenation (see~\Cref{fig:CompressionInit}). The collection $\mathcal{W}$ is naturally a 3rd-order tensor in $\mathbb{R}^{d \times p \times N}$; our long-axis concatenation corresponds to its mode-2 unfolding, and the resulting factorization into a shared $\mathbf{U}$ and layer-specific sparse $\mathbf{V}_i$ is the order-2 (matrix) case of a Tucker-1 decomposition with sparse core slices---distinct from higher-order methods such as tensor train or CP decomposition. Specifically, fine-grained parameter sharing across multiple layers can be achieved by expanding the projection matrix $\mathbf{V}$ and the shared basis $\mathbf{U}$. Sharing neurons between layers in this manner may be viewed as a factorization of $\mathcal{W}$, where the first factor $\mathbf{U}$ is shared across $N$ layers and the second, layer-specific factor $\mathbf{V}$ is sparse. Existing low-rank decomposition techniques can therefore be employed to obtain a shared basis---orthogonal at initialization via SVD, though not constrained to remain so during optimization---while sparsity in the projection matrices is induced using standard pruning and sparse training methods. When the parameter budget yields $r \le d$, this is a genuine low-rank factorization of $\mathcal{W}$ with an orthogonal basis $\mathbf{U}$ at initialization; at higher budgets, $r$ may exceed $d$, in which case $\mathbf{U}$ acts as an overcomplete shared dictionary that is no longer orthogonal, and parameter reduction arises from sharing and sparsity rather than rank reduction (discussed further in \S\ref{sec:method}). 

In the following sections, we investigate optimal layer-tying strategies within our framework, using a 12-block \textsc{DeiT-B} encoder with a single MLP module per block, pretrained on ImageNet-1k~\citep{ImageNet}. Specifically, we focus on MLP modules, which account for the majority of parameters (e.g., 70.5\% in \textsc{Gemma-2-9B}~\citep{Gemma2}) and comprise two fully connected (FC) layers (i.e., FC-1 and FC-2 for \textsc{DeiT-B}) of dimensions \(\mathbb{R}^{d\times p}\) and \(\mathbb{R}^{p\times d}\) with \(p=4d\). The ``parameter budget'' denotes the fraction of nonzero parameters retained after truncated SVD and sparsification; e.g., 25\% retains one-quarter of each MLP's weights. We measure the overall compression ratio as the percentage reduction in model size in bits, including sparsity metadata; e.g., 20\% compression reduces storage by 20\%. Concretely, the compressed model size (in bits) is
\begin{equation}\label{eq:compressed_size}
    |\theta_{\text{non-MLP}}| \cdot b \;+\; G \cdot |\mathbf{U}| \cdot b \;+\; \textstyle\sum_{i} \bigl[(1{-}s) \cdot |\mathbf{V}_i| \cdot b \;+\; |\mathbf{V}_i|\bigr],
\end{equation}
where $G$ is the number of groups, $s$ the sparsity level, $b$ the bits per parameter, and the final $|\mathbf{V}_i|$ term accounts for the 1-bit mask per element in unstructured sparsity. The compression ratio is then $1$ minus this quantity divided by the original model size $|\theta_{\text{original}}| \cdot b_{\text{original}}$. Note that non-MLP parameters (attention, embeddings, layer norms) remain uncompressed and are counted at full precision (full derivation in Appendix~\ref{app:compression_ratio}). The rank $r$ of the shared basis is not a free hyperparameter---given a parameter budget $b_{\%}$, the number of FC layers in the group $N$ (number of blocks $\times$ FCs per block), and target sparsity $s$, it is uniquely determined by:
\begin{equation}\label{eq:rank}
r = \left\lfloor \frac{b_{\%}\,N\,d\,p}{\,d + N(1{-}s)\,p\,} \right\rfloor.
\end{equation}

Conceptually, \gls{fips} differs from standard low-rank compression in three key aspects: 
(i) the basis is shared across layers rather than learned per layer; 
(ii) sparsity is imposed on the projection factors rather than the basis, enabling flexible neuron recombination; and 
(iii) optimization is performed jointly across layers using block-wise reconstruction loss. 
Together, these components transform low-rank approximation into a fine-grained parameter sharing mechanism.

\subsection{Optimal Sparsity Allocation Across Factors}\label{sec:OptimalDecomp}
\begin{figure}[!t]
    \centering
    \begin{subfigure}[t]{0.36\textwidth}
        \centering
        \includegraphics[width=\textwidth]{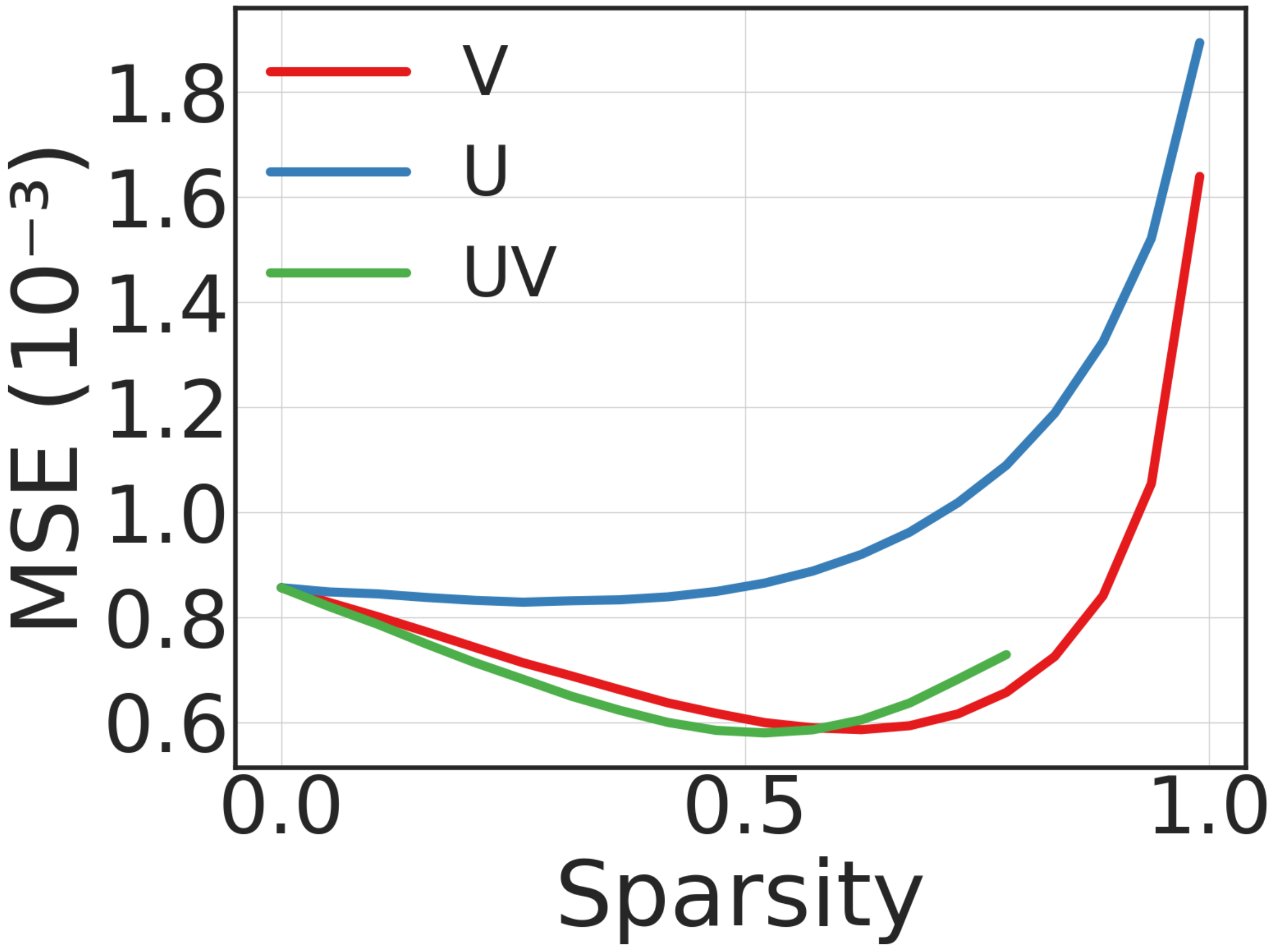}
        \caption{FC-1}
        \label{fig:MLP1_MSE}
    \end{subfigure}
    \hspace{0.1\textwidth}
    \begin{subfigure}[t]{0.36\textwidth}
        \centering
        \includegraphics[width=\textwidth]{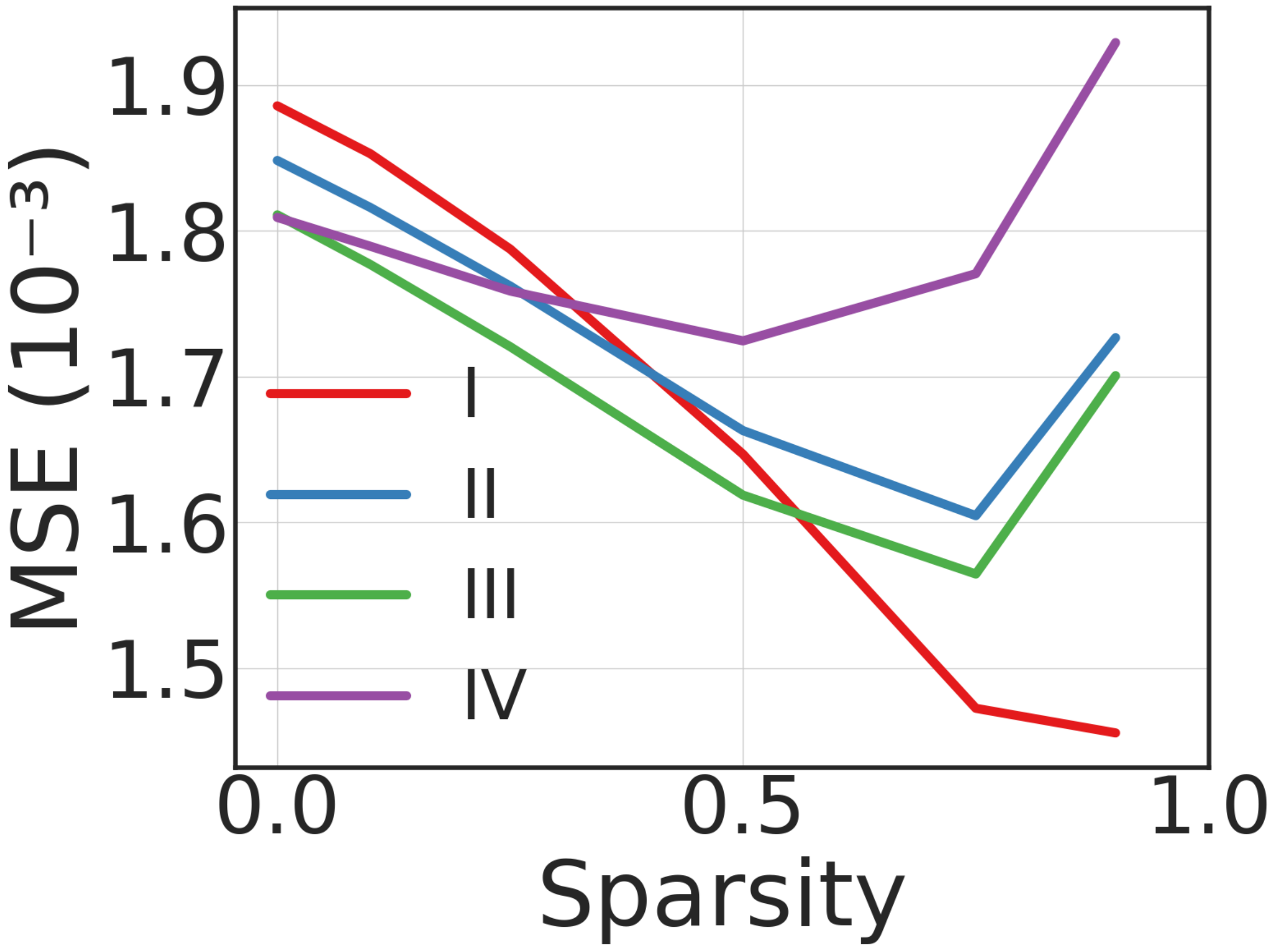}
        \caption{Sharing Strategies}
        \label{fig:concat_dims}
    \end{subfigure}
    \caption{\textbf{Initial Experiments.} (\subref{fig:MLP1_MSE}) Reconstruction error with varying levels of sparsity on different factors of the low-rank decomposition of FC-1 under 25\% parameter budget. Results are analogous for FC-2, i.e., inducing sparsity on the larger factor yields a higher rank and, thus, lower reconstruction error. (\subref{fig:concat_dims}) Mean reconstruction error across four FCs of two distinct encoder blocks' MLPs under various parameter sharing schemes and sparsities. See~\Cref{sec:WhichDimsToShare} for details.}
    \label{fig:motivation_how}
\end{figure}

Before implementing parameter sharing via shared bases, we decompose individual FC layers of MLP modules using truncated SVD at a 25\% parameter budget. Subsequently, sparsity is introduced by pruning low-magnitude values. Specifically, we examine sparsity induction in: (1) $\mathbf{U}$, (2) $\mathbf{V}$, and (3) both $\mathbf{U}$ and $\mathbf{V}$. Throughout this process, we vary the sparsity levels of the matrices while maintaining a constant total number of non-zero parameters. The resulting reconstruction errors are presented in~\Cref{fig:MLP1_MSE}. Our experiments show that the lowest reconstruction errors occur at sparsity levels between 60\% and 80\%, as confirmed by a sparsity sweep on ImageNet-1k with \textsc{DeiT-B} (\Cref{fig:sensitivity}), particularly when sparsity is imposed on the larger factor matrix $\mathbf{V}$. We attribute this to the higher redundancy in larger matrices, which facilitates more efficient pruning.

\subsection{Weight Concatenation and Sharing Dimensions}\label{sec:WhichDimsToShare}
We investigate parameter sharing across multiple layers by analyzing four FC layers drawn from two distinct MLP modules. To align their dimensions, we transpose the second FC layer of each module, representing every layer as $\mathbf{W}\in\mathbb{R}^{d\times 4d}$. We then examine four concatenation strategies for constructing a shared weight block $\mathbf{W}_s$:

\begin{enumerate}[label=(\Roman*),itemsep=4pt]
    \item \textbf{Long-axis concatenation.}  
    All four weight matrices $\mathbf{W}$ are concatenated along their output (neuron) dimension, resulting in $\mathbf{W}_s\in\mathbb{R}^{d\times 16d}$.

    \item \textbf{Module-wise long, inter-module short.}  
    Inside each MLP, its two FC layers are first concatenated along the output dimension, producing  $\mathbf{W}_s\in\mathbb{R}^{2d\times 8d}$.

    \item \textbf{Module-wise short, inter-module long.}  
    Inside each MLP, its two FC layers are first concatenated along the input dimension, again yielding  $\mathbf{W}_s\in\mathbb{R}^{2d\times 8d}$.

    \item \textbf{Short-axis concatenation.}  
    All four $\mathbf{W}$ matrices are concatenated along their input (feature) dimension, aligning features rather than neurons and resulting in  
    $\mathbf{W}_s\in\mathbb{R}^{4d\times 4d}$.
\end{enumerate}

For each concatenated block $\mathbf{W}_s$, we perform truncated SVD to retain the top $r$ singular vectors, followed by sparsification of the right singular matrix $\mathbf{V}$ via magnitude pruning~(see~\Cref{sec:OptimalDecomp}). Reconstruction is then obtained using the resulting shared basis, and mean squared error (MSE) is reported in~\Cref{fig:concat_dims}. Empirically, concatenation along the longer dimension consistently achieves the lowest reconstruction error—particularly under high sparsity—motivating our choice of full long-axis concatenation throughout. Further implementation details are provided in~\Cref{app:Concat}.

\subsection{Parameter Sharing Across Layers}\label{sec:LayerSharing}
\begin{figure*}[!t]
    \centering
    \begin{subfigure}[b]{0.56\textwidth}
        \centering
        \includegraphics[width=\textwidth]{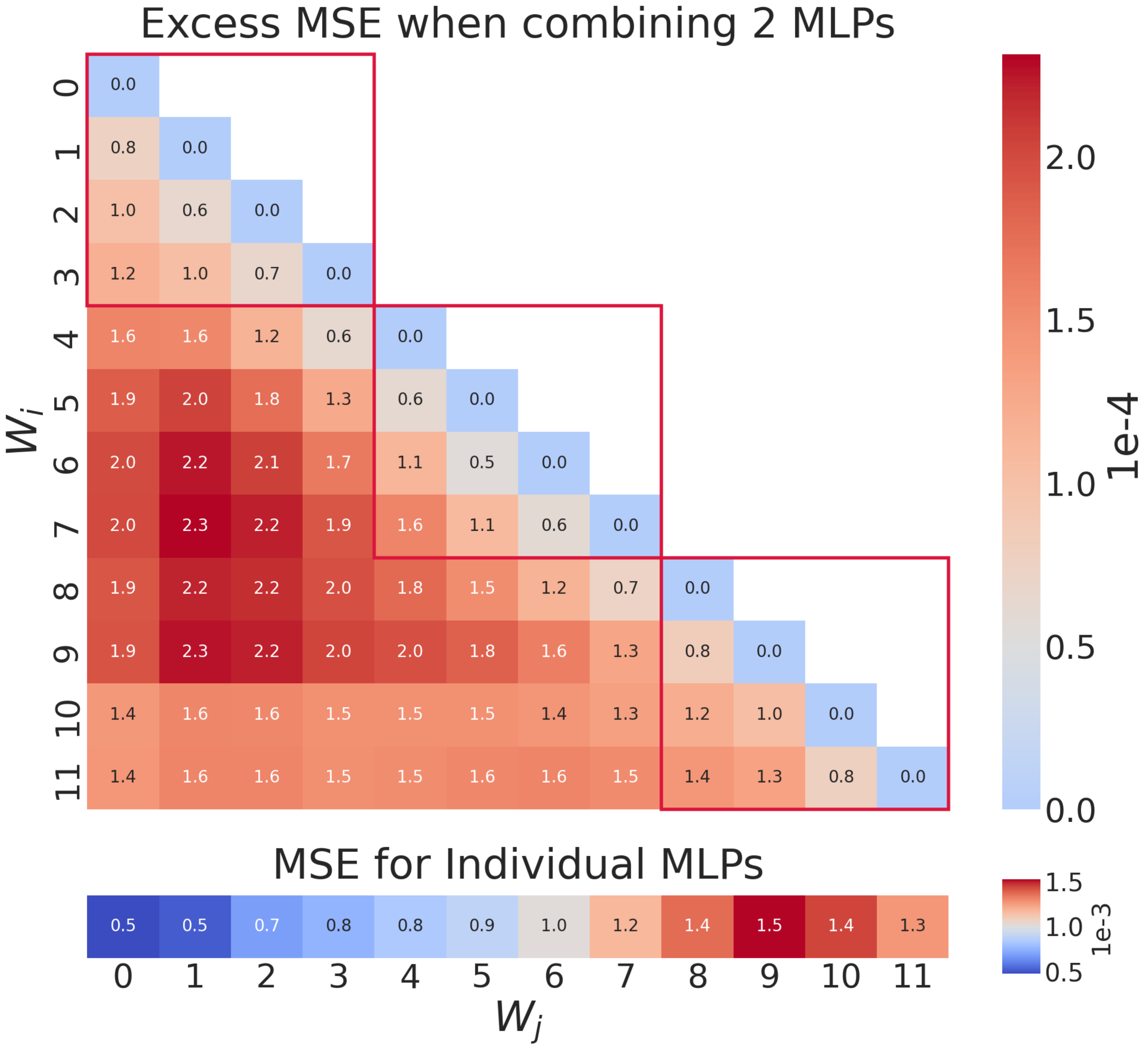}
        \caption{MSE when compressing MLP pairs.}
        \label{fig:2d_heat}
    \end{subfigure}
    \begin{minipage}[b]{0.42\textwidth}
        \centering
        \begin{subfigure}[b]{0.75\textwidth}
            \centering
            \includegraphics[width=\textwidth]{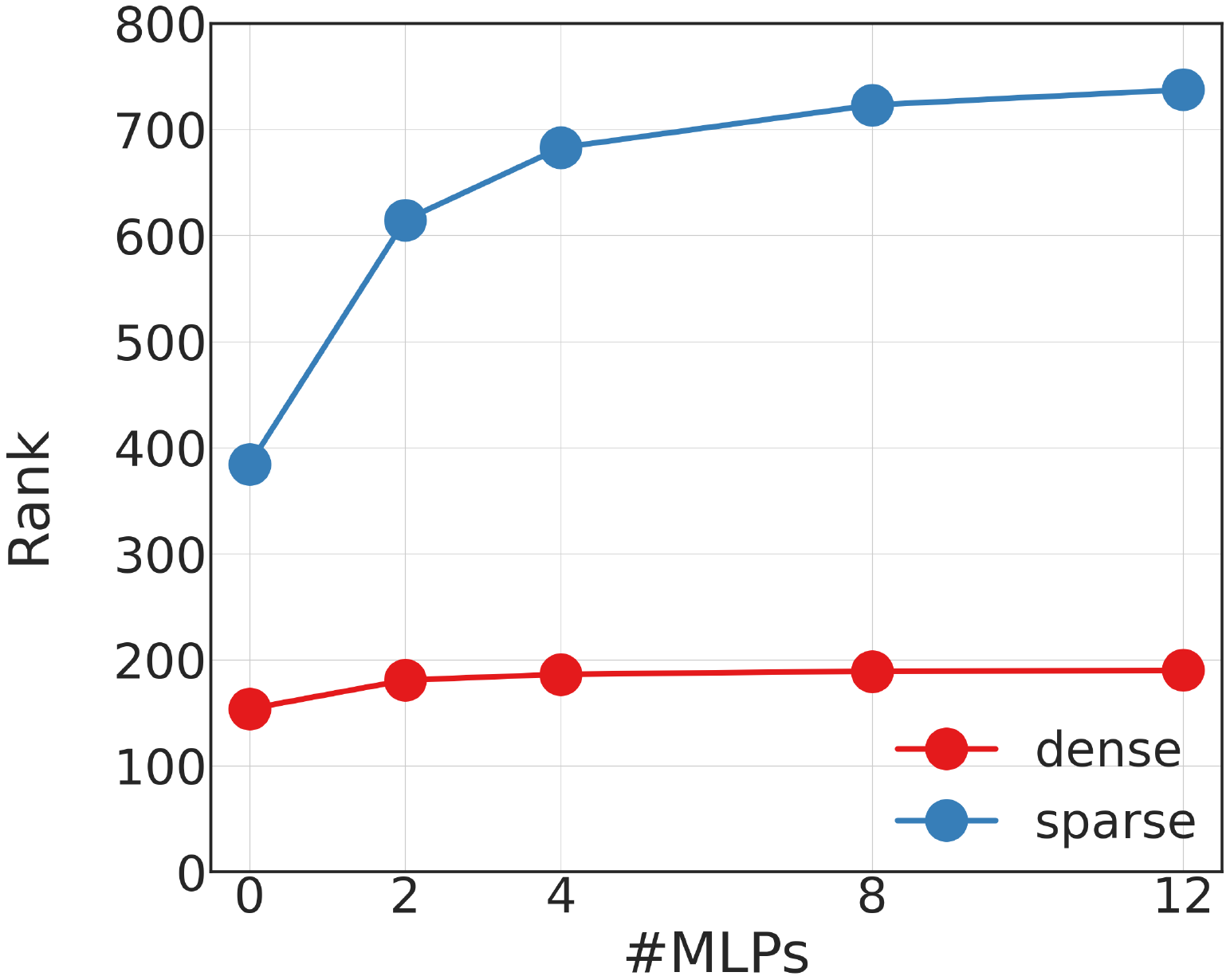}
            \caption{Rank vs.\ \#MLPs.}
            \label{fig:dense_vs_sparse_rank}
        \end{subfigure}
        
        \vspace{0.2cm}
        
        \begin{subfigure}[b]{0.75\textwidth}
            \centering
            \includegraphics[width=\textwidth]{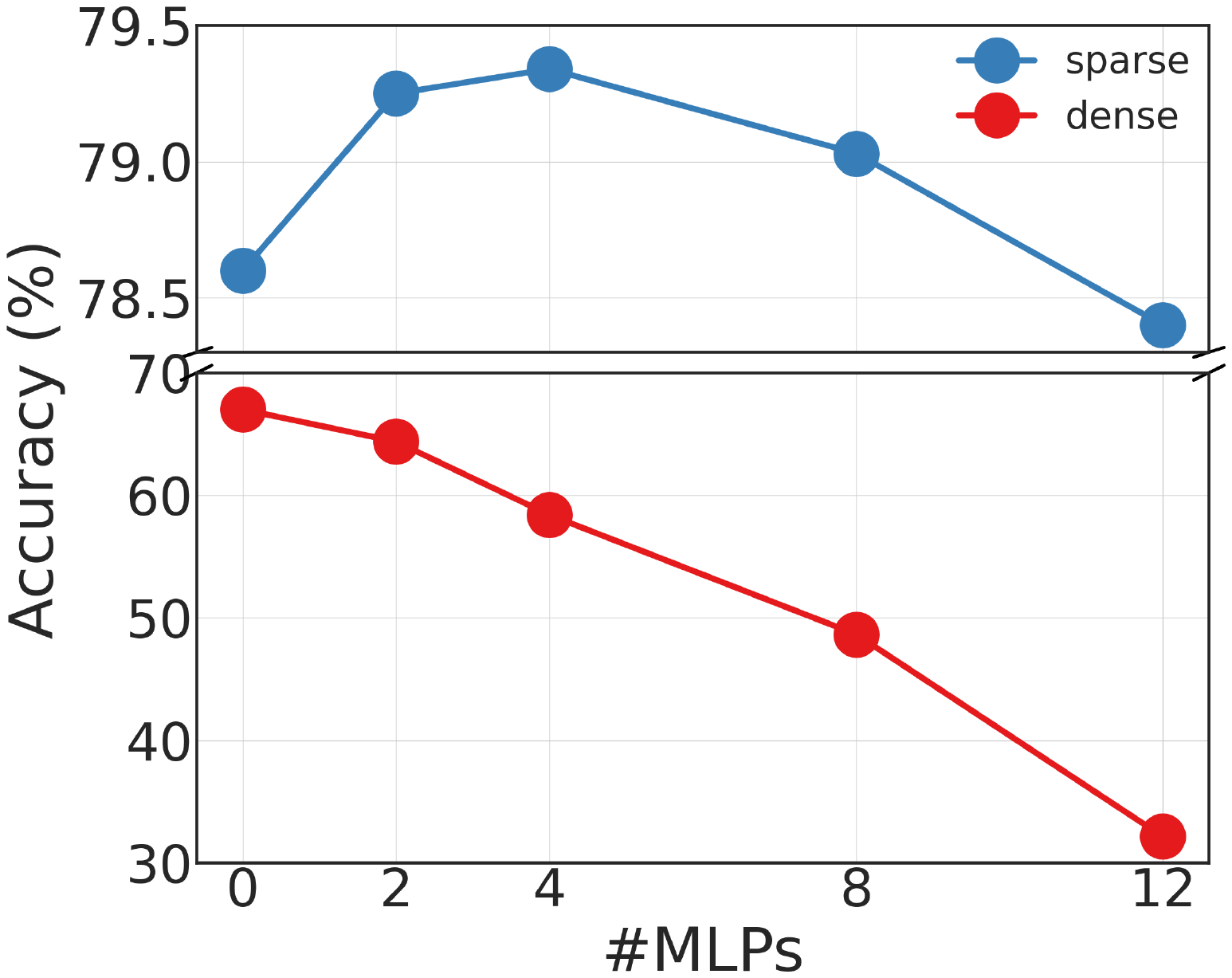}
            \caption{Accuracy vs.\ \#MLPs.}
            \label{fig:dense_vs_sparse_accuracy}
        \end{subfigure}
    \end{minipage}
    \caption{\textbf{Parameter Sharing Groups.}
    (\subref{fig:2d_heat} top) MSE increases when sharing $\mathbf{U}$ across different MLP modules; red squares over the diagonal indicate that sharing adjacent modules enhances reconstruction. 
    (\subref{fig:2d_heat} bottom) MSE for compressing individual MLP modules, showing that sharing $\mathbf{U}$ among consecutive layers typically results in the lowest error. 
    (\subref{fig:dense_vs_sparse_rank}) For a fixed parameter budget, the rank of the shared basis $\mathbf{U}$ stabilizes around four MLP modules, aligning with the optimal group size (\subref{fig:dense_vs_sparse_accuracy}) for maximizing accuracy in \textsc{DeiT-B}, while the dense counterpart continuously worsens.}
    \label{fig:motivation_where}
\end{figure*}

This section examines redundancy and interdependencies among MLP modules to identify optimal parameter sharing groupings. We first decompose each module individually at rank \(r=180\) and plot the resulting mean squared error in the lower panel of~\Cref{fig:2d_heat}. The error increases nearly monotonically with module depth, indicating that deeper layers require greater representational capacity. Next, we evaluate pairwise parameter sharing between modules \(i\) and \(j\) through a shared basis \(\mathbf{U}\). Parameter sharing reduces the total number of unique parameters but increases the reconstruction error for each module. We denote the average error increase due to parameter sharing between modules \(i\) and \(j\) as \(MSE^{\downarrow}_{i,j}\):
\[
  MSE^{\downarrow}_{i,j}
  = \frac{(MSE_{i,j} - MSE_{i}) + (MSE_{j,i} - MSE_{j})}{2},
\]
where \(MSE_{i,j}\) denotes the error of module $i$ when sharing a basis with module $j$. \Cref{fig:2d_heat} shows that adjacent modules exhibit the smallest error increase due to parameter sharing, motivating the practice of grouping consecutive layers.

We then explore the effect of grouping multiple MLP modules. Increasing group size allows a higher rank for the shared basis \(\mathbf{U}\), as shown in~\Cref{fig:dense_vs_sparse_rank}. This benefit is most pronounced when the projection matrices \(\mathbf{V}\) are sparsified. However, a higher rank does not always improve task performance, since \(\mathbf{U}\) must capture a larger set of neurons.~\Cref{fig:dense_vs_sparse_accuracy} demonstrates that sharing across four consecutive MLP modules yields the highest post-compression accuracy. 

Overall, our results reveal that (i) deeper modules require greater capacity when compressed in isolation—an effect we confirm with global pruning experiments in~\Cref{Sec:Ablation}; (ii) parameter sharing between adjacent layers curbs the rise in reconstruction error; and (iii) there exists an optimal group size that balances basis rank against sparsity. We encode this choice in the \emph{grouping hyperparameter} $\beta = [\beta_1, \ldots, \beta_G]$, an ordered list where $\beta_g$ gives the number of consecutive blocks whose MLP weights are tied in group~$g$, with $\sum_{g=1}^{G}\beta_g = L$ (total blocks). For \textsc{DeiT-B}, $\beta = [4, 4, 4]$ achieves the best trade-off; for LLMs a tapered grouping with smaller groups at the network extremes works best (e.g., $\beta = [4, 6, 6, 6, 6, 4]$ for \textsc{Llama-7B}). A full sweep is reported in~\Cref{app:block_groups}.

\section{Fine-grained Parameter Sharing}\label{sec:method}
The insights from~\Cref{sec:ParamSharing} motivate \gls{fips}, an efficient cross-block parameter sharing algorithm grounded in sparse tensor decomposition. \gls{fips} comprises three stages:

\begin{enumerate}[nolistsep]
    \item \textbf{Shared Initialization.} Tie the FC layers within each MLP group and apply truncated SVD to their concatenation (see~\Cref{fig:CompressionInit}), yielding a shared basis \(\mathbf{U}\) and projection matrices \(\{\mathbf{V}_i\}\).
    \item \textbf{Local Error Minimization.} Using a small calibration dataset \(D\)~(\Cref{sec:ViTExperiments}), optimize \(\mathbf{U}\) and each \(\mathbf{V}_i\) to minimize the \(\ell_2\) discrepancy between original and compressed activations while enforcing target sparsity in \(\mathbf{V}_i\).
    \item \textbf{Global Error Minimization (Optional).} Fine-tune the compressed model end-to-end under a dynamic sparse training regime to recover performance at higher compression ratios.
\end{enumerate}

\paragraph{Shared Initialization.}\label{par:shared-init}
We begin by compressing the pretrained model through parameter sharing, achieved by concatenating and decomposing multiple FC layers simultaneously. For higher parameter budgets and sparsity levels (e.g., 50\% and 75\%, respectively, for \textsc{DeiT-B}), the rank of the shared factor $\mathbf{U}$ can exceed the model dimension $d$. In this regime, $\mathbf{U}$ is no longer a low-rank basis but an overcomplete shared dictionary: parameter reduction comes from sharing $\mathbf{U}$ across blocks and sparsity in $\mathbf{V}$, not from rank reduction. The rank $r$ is determined by Equation~\ref{eq:rank} and remains fixed throughout optimization. To initialize the $k = r - d$ additional dimensions, we compare three strategies: (1)~random growth, initializing new neurons in $\mathbf{U}$ to zero and in $\mathbf{V}$ via~\citet{KaimingHeInit}; (2)~neuron splitting, duplicating and halving top neurons following Net2Net~\citep{Net2Net}; and (3)~hybrid initialization, setting new neurons in $\mathbf{U}$ to zero and deriving those in $\mathbf{V}$ from the top-$k$ singular vectors scaled by $1/\tau$, where $\tau > 1$ dampens their initial contribution so they are learned gradually~\citep{GradMax}. After sweeping $\tau$ (see~\Cref{app:growNeurons}), hybrid initialization outperforms the alternatives by 1--2 percentage points in top-1 accuracy.

Formally, the parameters of a group of FC layers across MLP modules, $\mathbf{W}_1, \mathbf{W}_2, \dots, \mathbf{W}_N$, are concatenated into $\mathcal{W} = [\mathbf{W}_1; \mathbf{W}_2; \dots; \mathbf{W}_N]$, where $\mathbf{W}_i \in \mathbb{R}^{d \times p}$.\footnote{The second FC layer is transposed to match the dimensions of the first.} We then apply truncated SVD, \(\mathcal{W} = \mathbf{U} \mathbf{\Sigma} \mathbf{\hat V}\), to obtain a low-rank approximation, where $\mathbf{U} \in \mathbb{R}^{d \times r}$, $\mathbf{\Sigma} \in \mathbb{R}^{r \times r}$, and $\mathbf{\hat V} \in \mathbb{R}^{r \times (N \cdot p)}$. The factor $\mathbf{U}$ is shared among all layers within the group and remains dense due to its small size. We then multiply $\mathbf{\hat V}$ by the singular values to obtain the projection matrix $\mathbf{V}= \mathbf{\Sigma}\mathbf{\hat V}$. Finally, the weights are reconstructed as $\mathbf{W^\prime}_i = \mathbf{U} \mathbf{V}_i$, where each $\mathbf{V}_i$ is the slice of $\mathbf{V}$ corresponding to weight matrix $\mathbf{W}_i$.

\paragraph{Local Error Minimization.} In the second phase of \gls{fips}, we compute the input and output activations of the original FC layers using a calibration dataset $D$ (described in \Cref{sec:ViTExperiments}). We use these activations to optimize the compressed layers by minimizing the \textit{$\ell_2$-loss} between the original and compressed activations:
\begin{equation}\label{eq:local_loss}
\operatorname*{argmin}_{\mathbf{U, V_i, \dots, V_N}} \sum_{i}^{N}{\| \mathbf{W}_{i}\mathbf{X}_{i} - \mathbf{U}\mathbf{V}_{i}\mathbf{X}_{i} \|_2^2},
\end{equation}where $\mathbf{X}_i$ is the inputs to the $i^{\textrm{th}}$ original FC layer. Sparsity on each $\mathbf{V}_i$ is enforced by the \textit{Sparsify} step of Algorithm~\ref{algorithm} rather than as an explicit constraint in the loss; $\mathbf{U}$ is left unconstrained, as orthogonality is only a byproduct of the SVD initialization and need not be preserved. 

We explore several sparse training and pruning strategies to identify a sparse \(\mathbf{V}\) during optimization: (a) \textit{Static Sparsity}, which fixes the sparsity pattern by retaining the top-magnitude connections before training~\citep{SparsityInDL}; (b) \textit{Gradual Magnitude Pruning (GMP)}~\citep{ToPruneOrNotToPrune}, which progressively increases sparsity by updating the mask every \(T\) steps according to the cubic schedule of~\citet{howtopruneLM}; and (c) \textit{RigL}~\citep{RigL}, which initializes as in (a) but updates sparse connectivity every \(\Delta T\) steps using both gradient and magnitude information. We adopt \textit{GMP} as our final strategy due to its superior performance, achieving up to 4\% higher top-1 accuracy on ImageNet-1k compared to the closest baseline. A detailed comparison of sparsification methods is given in~\Cref{tab:app:sparsity_ablation}. 

During this stage, parameters are shared across multiple MLP modules, allowing gradients to be computed one module at a time. As a result, optimization is significantly more resource-efficient than end-to-end fine-tuning.

\paragraph{Global Error Minimization.} In this optional stage, we fine-tune the learned parameter sharing scheme end-to-end to further improve performance and leverage the masks learned via GMP. Because the factors $\mathbf{V}_i$ are sparse, we employ the dynamic sparse training method \textit{RigL} during this stage, as it performs slightly better than \textit{Static Sparsity} (see~\Cref{Sec:RecoveryResults}).

\begin{algorithm}[tbp]
\caption{\acrlong{fips}}\label{algorithm}
\begin{algorithmic}[1]
\Require MLP parameters $\mathbf{W}_1,\cdots,\mathbf{W}_N \in \mathbb{R}^{d\times p}$, MLP inputs $\mathbf{A}_i$ and MLP function $\mathbf{f}(\mathbf{W}_i, \ \mathbf{A}_i)$, Target rank $r$, Learning Rate $\mathbf{\eta}$,  Steps $\mathbf{T}$. 
\State $\mathbf{U},\ \left[\mathbf{V}_1, \mathbf{V}_2, \cdots, \mathbf{V}_N \right] \gets \mathit{TruncatedSVD} \left( \left[\mathbf{W}_1; \mathbf{W}_2; \cdots; \mathbf{W}_N \right],\ k=r \right)$ 
\For{each training iteration $t = 1$ to $T$}
    \State  $\mathbf{G_U} = 0$ \Comment{Gradient accumulator for $\mathbf{U}$}
    \For{each block $i$}
        \State $\mathbf{V}_i \gets \mathit{Sparsify}(\mathbf{V}_i, \ t)$ \Comment{Potentially increase or adjust sparsity}
        \State $L_i \gets \mathit{MSE\_loss}(\mathbf{f}(\mathbf{W}_i,\ \mathbf{A}_i), \quad \mathbf{f}(\mathbf{U}\,\mathbf{V_i}, \ \mathbf{A}_i))$
        \State $\mathbf{V}_i \gets \mathbf{V}_i - \eta \nabla_{\mathbf{V}_i} L_i$
        \State $\mathbf{G_U} \gets \mathbf{G_U} + \mathbf{\nabla_{U}} L_i$
    \EndFor
    \State $\mathbf{U} \gets \mathbf{U} - \dfrac{\eta}{N} \mathbf{G_U}$
\EndFor
\State \Return $\mathbf{U}, \  [\mathbf{V}_1, \ldots, \mathbf{V}_N]$
\end{algorithmic}
\end{algorithm}

\section{Main Results}\label{sec:MainResults}
\paragraph{Evaluation Metrics.} For ViTs we report top-1 classification accuracy on ImageNet-1k validation. For LLMs we report perplexity (PPL), defined as $\text{PPL} = \exp\!\bigl(-\tfrac{1}{T}\sum_{t=1}^{T}\log p(x_t \mid x_{<t})\bigr)$, on WikiText-2 and C4, as well as downstream task accuracy via LM-Evaluation-Harness~\citep{LMEvalEleuther}. During compression, reconstruction quality is measured by mean squared error (MSE) between original and compressed activations, i.e., the per-layer $\ell_2$ loss in \Cref{eq:local_loss}. Lower PPL and MSE indicate better quality; higher accuracy is better.

\subsection{Vision Transformers}\label{sec:ViTExperiments}
\paragraph{Experimental Setup.} We evaluate \gls{fips} on \textsc{DeiT-B}~\citep{DeiT} and \textsc{Swin-L}~\citep{SWIN}. Each model is calibrated on 2{,}560 ImageNet-1k images for 20 epochs, sufficient for convergence; calibration completes in under one hour on a single NVIDIA~A6000—comparable to or cheaper than existing post-training compression methods, none of which report full training cost breakdowns. For parameter sharing, we group every four MLP modules in \textsc{DeiT-B}. In \textsc{Swin-L}, which consists of four stages with 2, 2, 18, and 2 encoder blocks, we share across entire stages for the three smaller ones and use groups of six blocks in the larger stage. Additional hyperparameters are provided in~\Cref{app:hyperParams}; sensitivity analyses are in \Cref{fig:sensitivity}.

\paragraph{ImageNet-1k.}\label{Sec:RecoveryResults}
We compare \gls{fips} to the two closest ViT compression baselines that share its post-training, factorization-based setting without distillation or LoRA: Adaptive Atomic Feature Mimicking (AAFM), which compresses output activations rather than weights, and Global Feature Mimicking (GFM), which fine-tunes the compressed network~\citep{CompressingTransformers}. Both \gls{fips} and \gls{fips}+FT match AAFM and GFM in compute and memory budgets. At a 40\% parameter budget, \gls{fips} outperforms AAFM by 1.36 points and GFM by 0.41 points despite GFM's higher cost (\Cref{tab:compression_methods_comparison_models}). This pattern holds across all budgets: \gls{fips} consistently achieves the highest accuracy while matching or improving upon the compute and memory budgets of AAFM and GFM. AAFM requires no fine-tuning; GFM requires full end-to-end fine-tuning; \gls{fips}'s local error minimization is strictly cheaper than GFM and comparable to AAFM. Notably, a 10\% parameter budget corresponds to roughly a 50\% compression ratio, where \gls{fips}+FT yields particularly strong gains. 

\paragraph{Transfer Learning.}
For transfer learning, we fine-tune for 100~epochs on CIFAR-100, Flowers102, Oxford-III-Pets, and iNaturalist~2019~\citep{CIFAR100Citation,FlowersCitation,PetsDataset,iNaturalistCitation}, following~\citet{CompressingTransformers}. We use \textit{AdamW}~\citep{AdamW} with learning rates selected from 12 log-spaced values. Models compressed with \gls{fips} transfer significantly better, as shown in~\Cref{tab:CompTransformTransfer}.

\begin{table}[tbp]
    \centering
    \caption{\textbf{ViT Compression Results.} ImageNet-1k top-1 validation accuracy of \textsc{DeiT-B} (81.85\%)~\citep{DeiT} and \textsc{Swin-L} (86.24\%)~\citep{SWIN} compressed using \gls{fips} and AAFM/GFM across MLP parameter budgets, comparing layer-wise (\gls{fips}) and global error minimization (\gls{fips}+FT). Parameter budgets refer to the fraction of MLP parameters retained (see \S\ref{sec:ParamSharing}); the corresponding whole-model compression ratios (via Equation~\ref{eq:compressed_size}) are shown in the second row. AAFM/GFM\textsuperscript{\dag} results are from~\citet{CompressingTransformers}; -- denotes unreported metrics.}
    \label{tab:compression_methods_comparison_models}
    \begin{tabular}{@{}lp{2em}p{2.1em}p{2em}p{2em}p{2em}p{2em}p{2em}p{2em}p{2em}p{2em}@{}}
    \toprule
    Parameter Budget & \multicolumn{2}{c}{10\%} & \multicolumn{2}{c}{25\%} & \multicolumn{2}{c}{40\%} & \multicolumn{2}{c}{50\%} & \multicolumn{2}{c}{75\%} \\
    {\small(Comp.\ Ratio)} & \multicolumn{2}{c}{\small(${\sim}$57\%)} & \multicolumn{2}{c}{\small(${\sim}$45\%)} & \multicolumn{2}{c}{\small(${\sim}$33\%)} & \multicolumn{2}{c}{\small(${\sim}$25\%)} & \multicolumn{2}{c}{\small(${\sim}$6\%)} \\
    \cmidrule(l){2-3}\cmidrule(l){4-5}\cmidrule(l){6-7}\cmidrule(l){8-9}\cmidrule(l){10-11}
    Method / Model & \textsc{DeiT} & \textsc{Swin} & \textsc{DeiT} & \textsc{Swin} & \textsc{DeiT} & \textsc{Swin} & \textsc{DeiT} & \textsc{Swin} & \textsc{DeiT} & \textsc{Swin} \\
    \midrule
    \text{AAFM~\textsuperscript{\dag}} & -- & -- & -- & -- & 80.33 & -- & 81.21 & 85.04 & 81.76  & 85.94 \\
    \text{GFM~\textsuperscript{\dag}} & -- & -- & -- & -- & 81.28 & -- & 81.62 & 85.44 & 81.83 & 86.01 \\
    \midrule
    \text{\gls{fips}} (ours) & 70.04 & 74.04 & 80.64 & 84.78 & \textbf{81.69} & \textbf{85.69} & \textbf{81.83} & \textbf{85.99} & \textbf{81.82} & 86.21 \\
    \text{\gls{fips}} + FT (ours) & \textbf{77.26} & \textbf{82.13} & \textbf{81.31} & \textbf{85.16} & 81.54 & 85.68 & 81.54 & \textbf{85.99} &  \textbf{81.82} & \textbf{86.22} \\
    \bottomrule
    \end{tabular}
\end{table}

\paragraph{Latency and Memory Profiling.}
Structured sparsity patterns enable efficient hardware implementations with minimal quality impact, as demonstrated in~\Cref{tab:structured_sparsity} using \textit{NMGMP}. With 2:4 structured \gls{fips}, the accuracy degradation remains just above 1\% at 10\% and 25\% MLP parameter budgets; for all other settings, the impact is negligible. We further evaluate \gls{fips} with alternative structured sparsity pruners—\textit{STE}, \textit{SR-STE}~\citep{SRSTE}, and \textit{NMSRigL}~\citep{JaxPruner, SRigL}—in~\Cref{tab:app:structured_sparsity}. Latency and memory profiling with the optimal \textit{NMGMP}+\gls{fips} setup leverages NVIDIA's tensor core support for 2:4 sparsity~\citep{mishra_accelerating_2021} on GPUs and Neural Magic's DeepSparse Engine~\citep{neural_magic_neuralmagicdeepsparse_2021} on CPUs, as shown in~\Cref{fig:latency-benchmarks} and detailed in~\Cref{app:latency}.

\begin{table}[tbp]
  \centering
    \caption{(\subref{tab:CompTransformTransfer}) Top-1 accuracy of \textsc{DeiT-B} and compressed variants using GFM and \gls{fips} under different parameter budgets. GFM\textsuperscript{\dag} and Original\textsuperscript{\dag} results are from \citet{CompressingTransformers} and \citet{DeiT}. (\subref{tab:structured_sparsity}) ImageNet-1k top-1 accuracy of \textsc{DeiT-B} with \gls{fips} at 2:4 structured sparsity (N:M GMP~\citep{JaxPruner}) versus 50\% unstructured sparsity.}
  \label{tab:combined}
  \begin{subtable}[t]{0.55\textwidth}
    \centering
    \caption{}
    \label{tab:CompTransformTransfer}
    \small
    \setlength{\tabcolsep}{2.5pt}
    \renewcommand{\arraystretch}{1}
    \begin{tabular}{@{}lccccccc@{}}
      \toprule
      & Original\textsuperscript{\dag}
      & \multicolumn{2}{c}{GFM\textsuperscript{\dag}}
      & \multicolumn{3}{c}{\text{\gls{fips}}+RigL FT (ours)} \\
      \cmidrule(lr){2-2} \cmidrule(lr){3-4} \cmidrule(lr){5-7}
      P.\ Budget & 100 & 40\% & 50\% & 25\% & 40\% & 50\% \\
      {\scriptsize(Comp.\ Ratio)} & {\scriptsize --} & {\scriptsize(${\sim}$33\%)} & {\scriptsize(${\sim}$25\%)} & {\scriptsize(${\sim}$45\%)} & {\scriptsize(${\sim}$33\%)} & {\scriptsize(${\sim}$25\%)} \\
      \midrule
      CIFAR-100        & 90.99 & 90.17 & 90.67 & 90.88 & 91.24 & \textbf{91.33} \\
      Pets             & \textbf{94.74} & 93.95 & 94.22 & 94.19 & 94.52 & 94.41 \\
      Flowers102       & 97.77 & 97.02 & 97.45 & 97.84 & 98.14 & \textbf{98.37} \\
      iNaturalist 2019 & 77.39 & 77.13 & 77.56 & 77.26 & 77.58 & \textbf{77.69} \\
      \bottomrule
    \end{tabular}
  \end{subtable}
  \hfill
  \begin{subtable}[t]{0.40\textwidth}
    \centering
    \caption{}
    \label{tab:structured_sparsity}
    \small
    \setlength{\tabcolsep}{2pt}
    \renewcommand{\arraystretch}{1.2}
    \begin{tabular}{@{}lccccc@{}}
      \toprule
      P.\ Budget & 10\% & 25\% & 40\% & 50\% & 75\% \\
      {\scriptsize(Comp.\ Ratio)} & {\scriptsize(${\sim}$57\%)} & {\scriptsize(${\sim}$45\%)} & {\scriptsize(${\sim}$33\%)} & {\scriptsize(${\sim}$25\%)} & {\scriptsize(${\sim}$6\%)} \\
      \midrule
      $2{:}4$ \gls{fips}   & 52.36 & 76.88 & 80.59 & 81.31 & 81.51 \\
       \gls{fips}       & 54.00 & 77.56 & 80.94 & 81.63 & 81.77 \\
      \bottomrule
    \end{tabular}
  \end{subtable}
\end{table}

These results demonstrate that \gls{fips} with structured sparsity translates compression into measurable inference gains on ViTs. Specifically, with a batch size of 64, \gls{fips} with 2:4 sparsity at a 22.14\% parameter budget yields a $1.31\times$ speedup on the NVIDIA A4000 and reduces peak VRAM allocation to approximately $0.79\times$ of the original requirement during \textsc{DeiT-B} inference. These gains generalize architecturally: \gls{fips} replaces every dense FC layer $\mathbf{W}\in\mathbb{R}^{d\times p}$ with the product $\mathbf{U}\mathbf{V}$, changing the forward pass from a single dense matrix--vector product ($2dp$ FLOPs per token) to a sparse product followed by a dense one ($2rp(1{-}s) + 2dr$ FLOPs per token, where $s$ is the sparsity level). This FLOP reduction depends only on the factorization parameters $(r, s, d, p)$, not on whether the FC layer resides in a ViT encoder or an LLM decoder. Consequently, the latency and memory improvements measured on \textsc{DeiT-B} are expected to transfer to LLM architectures at matching compression configurations, subject to hardware-specific kernel availability for the relevant dimensions.

\subsection{Large Language Models}\label{sec:LLMExperiments}
\paragraph{Experimental Setup.}
We evaluate \gls{fips} on three publicly available pretrained LLMs: \textsc{Llama-7B}~\citep{Llama1}, \textsc{Llama-3.1-8B}~\citep{Llama3}, and the instruction-tuned \textsc{Gemma-2-2B}~\citep{Gemma2}. \textsc{Gemma-2-2B} is included solely to demonstrate that \gls{fips} is orthogonal to QAT, while the other models are compared against the baselines described below. Unlike ViTs, these LLMs feature three FC layers per MLP in their decoder blocks. For parameter sharing, all FC layers from MLPs within the same group are concatenated. The number of MLPs per group is treated as a hyperparameter and detailed in~\Cref{app:block_groups}. Calibration and optimization follow the ViT protocol (described in~\Cref{sec:ViTExperiments}): activations are collected from \(8{,}192 \times 20\) SlimPajamas~\citep{SlimRedPajama} tokens, and block-wise error minimization is run for 40 epochs—without global error minimization (i.e., no full FT)—on a single NVIDIA A100 (80GB), completing within 10 hours per model on a single GPU. To our knowledge, none of the compared baselines report end-to-end compression time; \gls{fips} is the only method in this comparison that provides explicit cost figures.

\begin{table}[tbp]
  \centering
  \setlength{\tabcolsep}{3pt}
  \resizebox{\textwidth}{!}{%
  \begin{tabular}{@{}llccccccccccc@{}}
    \toprule
     & Method & WikiText-2$\downarrow$ & C4$\downarrow$ & Openb. & ARC-e & WinoG. & HellaS. &
    PIQA & MathQA & \textbf{Avg.}$\uparrow$ & TruthfulQA$\uparrow$ & GSM8K$\uparrow$  \\
    \cmidrule(l{3pt}r{3pt}){2-2}
    \cmidrule(l{3pt}r{3pt}){3-4}
    \cmidrule(l{3pt}r{3pt}){5-11}
    \cmidrule(l{3pt}r{3pt}){12-13}
    \multirow{6}{*}{\textsc{Llama-7B}}
          & \textcolor{gray}{Original} & \textcolor{gray}{5.68} & \textcolor{gray}{7.34} & \textcolor{gray}{0.34} & \textcolor{gray}{0.75} & \textcolor{gray}{0.70} & \textcolor{gray}{0.57} & \textcolor{gray}{0.79} & \textcolor{gray}{0.27} & \textcolor{gray}{0.57} & \textcolor{gray}{0.30} & \textcolor{gray}{0.09} \\
      \cmidrule(lr){2-13}
      & ASVD       & 11.14 & 15.93 & 0.29 & 0.53 & 0.64 & 0.41 & 0.68 & 0.17 & 0.45 & 0.21 & 0.04 \\
      & SVD-LLM    & 7.94  & 15.84 & 0.31 & 0.71 & 0.68 & 0.49 & 0.71 & 0.22 & 0.52 & 0.24 & 0.06 \\
      & SVD-LLM V2 & 7.12  & 10.47 & 0.32 & 0.72 & 0.70 & 0.52 & 0.75 & 0.24 & 0.54 & \textbf{0.27} & \textbf{0.07} \\
      & Basis Sharing & 7.74 & 15.03 & 0.28 & 0.66 & 0.66 & 0.46 & 0.71 & 0.25 & 0.50 & -- & -- \\
      \cmidrule(lr){2-13}
      & \gls{fips} (ours) & \textbf{6.06} & \textbf{8.10} & \textbf{0.32} & \textbf{0.72} &
                    \textbf{0.70} & \textbf{0.56} & \textbf{0.78} & \textbf{0.26} &
                    \textbf{0.56} & \textbf{0.27} & \textbf{0.07} \\
    \midrule
    \multirow{5}{*}{\textsc{Llama-3.1-8B}}
          & \textcolor{gray}{Original}
          & \textcolor{gray}{6.14} & \textcolor{gray}{9.47}
          & \textcolor{gray}{0.35} & \textcolor{gray}{0.80}
          & \textcolor{gray}{0.73} & \textcolor{gray}{0.60}
          & \textcolor{gray}{0.80} & \textcolor{gray}{0.40}
          & \textcolor{gray}{0.61} & \textcolor{gray}{0.49}
          & \textcolor{gray}{0.45} \\
      \cmidrule(lr){2-13}
      & ASVD       & 17.55 & 28.41 & 0.20 & 0.59 & 0.61 & 0.41 & 0.69 & 0.30 & 0.47 & 0.37 & 0.28 \\
      & SVD-LLM    & 11.82 & 20.05 & 0.29 & 0.77 & 0.64 & 0.51 & 0.72 & 0.30 & 0.54 & 0.45 & 0.31 \\
      & SVD-LLM V2 & 8.01  & 11.72 & \textbf{0.33} & \textbf{0.79} & 0.70 & 0.58 & 0.77 & 0.36 & 0.59 & \textbf{0.46} & 0.40 \\
      \cmidrule(lr){2-13}
      & \gls{fips} (ours) & \textbf{6.88} & \textbf{10.78} & \textbf{0.33} & \textbf{0.79} &
                    \textbf{0.72} & \textbf{0.59} & \textbf{0.78} & \textbf{0.38} &
                    \textbf{0.60} & \textbf{0.46} & \textbf{0.42} \\
    \bottomrule
  \end{tabular}}
    \caption{\textbf{LLM Compression Results.}
    \textsc{Llama-7B} and \textsc{Llama-3.1-8B} at 20\% compression. PPL$\downarrow$: perplexity on WikiText-2 and C4; Avg.$\uparrow$: mean over six classification tasks; TruthfulQA (BLEU$\uparrow$) and GSM8K (exact match$\uparrow$). ASVD from~\citet{ASVD}; SVD-LLM from~\citet{SVDLLM}; SVD-LLM V2 from~\citet{SVDLLMv2}; Basis Sharing from~\citet{BasisSharing}. All results follow~\citet{LMEvalEleuther}; -- indicates unreported metrics.}
    \label{tab:LLMResultsHarnessWide}
\end{table}

\paragraph{Baselines.}
We restrict comparisons to methods that share \gls{fips}'s regime: post-training, factorization-based MLP compression without LoRA or distillation, so that differences reflect the factorization strategy rather than auxiliary training signals. We benchmark against three SVD-based LLM compression methods: ASVD~\citep{ASVD}, which scales weights by activation statistics to mitigate outliers; SVD-LLM~\citep{SVDLLM}, which employs truncation-aware whitening; and SVD-LLM V2~\citep{SVDLLMv2}, which adds layer-wise rank allocation. All baseline metrics are from the original publications; we exclude LoRA~\citep{hu2021loralowrankadaptationlarge} enhancements for fair comparison. Moreover, we compare \gls{fips} against Basis Sharing~\citep{BasisSharing}, which also pursues cross-layer parameter sharing but relies on dense coefficients. In contrast, \gls{fips} enforces sparsity, which our ablations show to be essential (see~\Cref{tab:app:sparsity_ablation}).

\paragraph{Evaluation.}
We report perplexity on WikiText-2~\citep{WikiText2} and C4~\citep{C4Dataset}, six classification benchmarks, and two generation tasks (TruthfulQA, GSM8K) via LM-Evaluation-Harness~\citep{LMEvalEleuther}. At 20\% compression (\Cref{tab:LLMResultsHarnessWide}), \gls{fips} achieves the lowest perplexity on both WikiText-2 and C4 for \textsc{Llama-7B} and \textsc{Llama-3.1-8B}, while matching or outperforming all baselines on downstream tasks. Results at 40\% compression are provided in~\Cref{app:tab:llama7b_wikitext2}.

\setlength{\intextsep}{0pt}
\begin{wraptable}[13]{r}{0.42\textwidth}
\centering
\captionsetup{skip=3pt}
\label{tab:qat_table}
\small
\begin{tabular}{lccc}
    \toprule
    \textbf{Variant} & \textbf{Prec.} & \textbf{Comp.} & \textbf{PPL} $\downarrow$ \\
    \midrule
    Baseline   & BF16  & 1.0$\times$ & 15.61 \\
    QAT        & INT4  & 4.0$\times$ & 16.86 \\
    QAT        & INT2  & 8.0$\times$ & 41.86 \\
    FiPS       & BF16  & $1.5\times$ & 32.01 \\
    FiPS+QAT   & INT3  & $8.0\times$ & 35.43 \\
    \bottomrule
\end{tabular}
\caption{\textbf{QAT Results.} 3-bit QAT with \gls{fips} achieves 8$\times$ compression at lower PPL than 2-bit QAT alone.}
\end{wraptable}
\paragraph{Quantization-Aware Training (QAT).}
To assess \gls{fips} under low-precision regimes, we apply QAT to \textsc{Gemma-2-2B}. \Cref{tab:qat_table} reports WikiText-2 perplexity for 4-bit and 2-bit QAT ($4\times$ and $8\times$ compression) as well as \gls{fips} combined with 3-bit QAT. While 4-bit QAT matches \texttt{bfloat16} performance, 2-bit QAT degrades severely (PPL\,=\,41.86). Combining 3-bit QAT with \gls{fips} achieves the same $8\times$ compression as 2-bit QAT but with substantially lower perplexity (35.43 vs.\ 41.86). Both approaches degrade substantially from the \texttt{bfloat16} baseline; the key finding is that \gls{fips} recovers quality relative to aggressive quantization at the same compression factor.

\section{Ablations}\label{Sec:Ablation}
We examine the importance of various components of the \gls{fips} algorithm when compressing \textsc{DeiT-B} at 25\% parameter budget. We ablate the following key components:
\begin{enumerate}[noitemsep, topsep=2pt, partopsep=0pt, parsep=0pt]
    \item \textbf{Random Initialization (RI):} Using RI instead of SVD initialization results in a 1 percentage point drop in accuracy. 
    \item \textbf{Global Pruning (GP):} Using GP when sparsifying the factors $\mathbf{V}$ yields a 0.4 percentage point improvement over local pruning (LP), which enforces the same sparsity level for each group.
    \item \textbf{Scaling Vectors (SV):} Following \citet{DoRA}, FC weights are normalized, and the magnitudes initialize the SV for neuron scaling. This enhances LP but is less effective than GP.
\end{enumerate}

\begin{figure}[tbp]
    \centering
    \begin{subfigure}[b]{0.35\columnwidth}
        \centering
        \includegraphics[width=\textwidth]{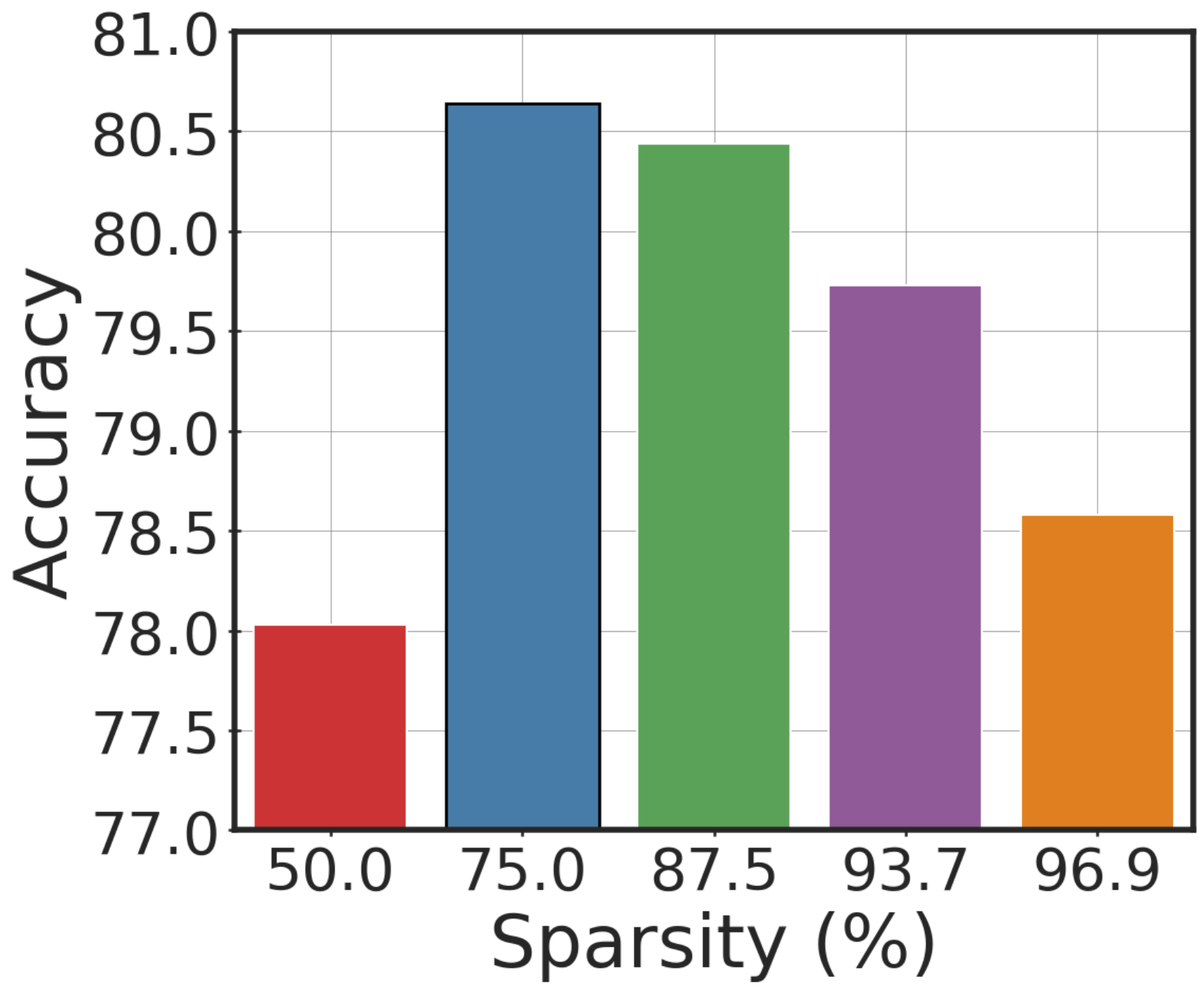}
        \caption{Optimal Sparsity.}
        \label{fig:sparsity_sweep}
    \end{subfigure}
    \hspace{0.1\columnwidth}
    \begin{subfigure}[b]{0.35\columnwidth}
        \centering
        \includegraphics[width=\textwidth]{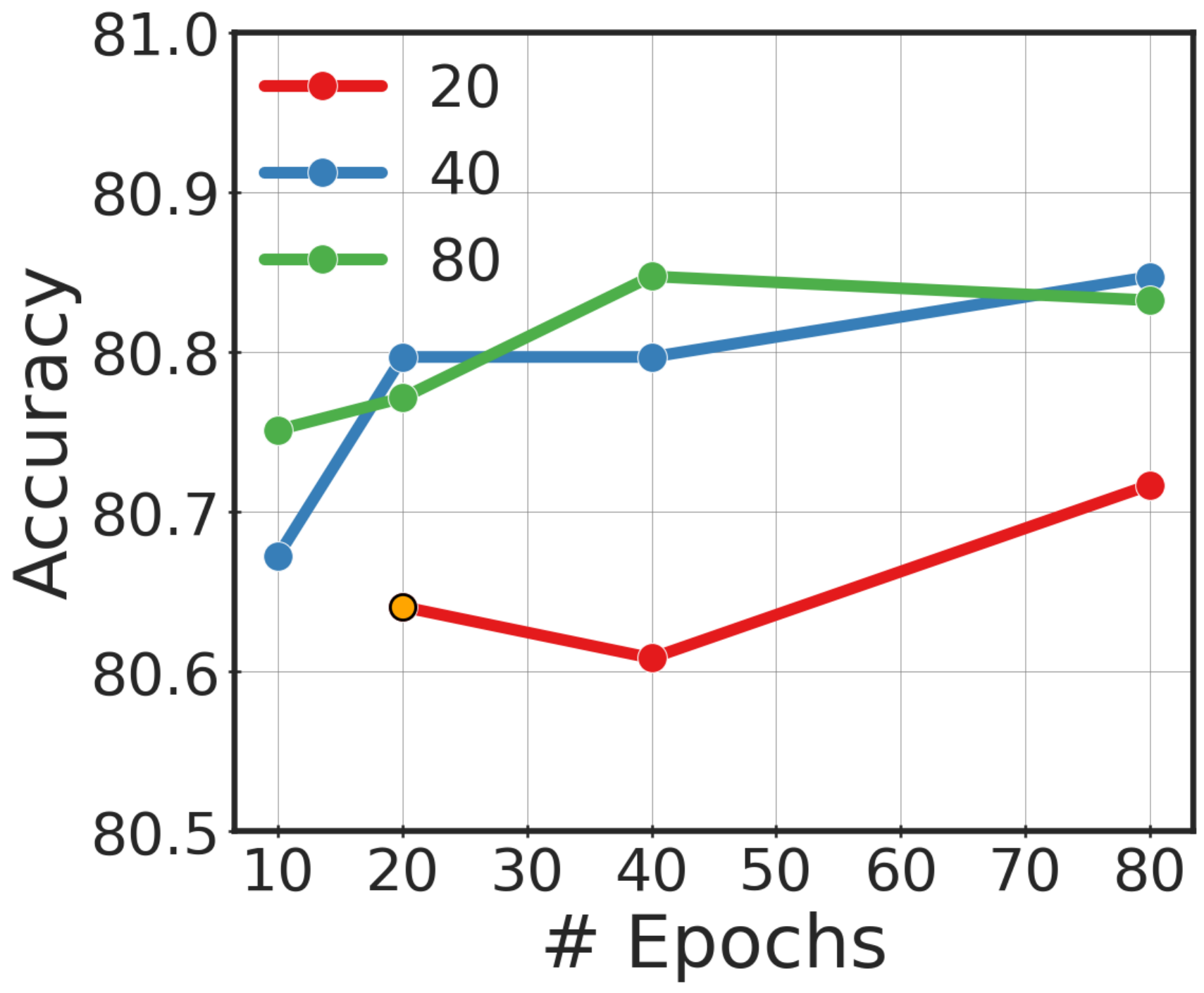}
        \caption{Calibration Data.}
        \label{fig:epoch_batch_sweep}
    \end{subfigure}
    \caption{\textbf{Sensitivity Analysis.} (a) Impact of sparsity levels on \textsc{DeiT-B} accuracy; 75\% is optimal. (b) Effect of calibration data volume and training duration; 20 epochs over 20 batches suffices.}
    \label{fig:sensitivity}
\end{figure}

An analysis on sparsity, methods, and calibration settings using \textsc{DeiT-B} confirms that GMP with 75\% sparsity and 20 epochs over 20 batches yields optimal performance (\Cref{fig:sensitivity}). Figure~\ref{fig:MLP1_MSE} shows the rank--sparsity trade-off at a fixed 25\% parameter budget: as sparsity on $\mathbf{V}$ increases from 0\% to ${\sim}$80\%, the rank $r$ grows to compensate, and reconstruction error decreases until diminishing returns set in beyond 80\% sparsity. \Cref{tab:app:sparsity_ablation} compares all sparsification strategies across budgets for both \textsc{DeiT-B} and \textsc{Swin-L}: GMP consistently achieves the highest accuracy, while the dense baseline (no sparsity on $\mathbf{V}$) collapses at low budgets (e.g., 15.35\% at 10\%), underscoring that sparsity is not merely helpful but essential. \Cref{fig:sensitivity} further shows that 75\% sparsity is optimal and that performance is robust to calibration data volume beyond 20 batches.

\begin{table}[tbp]
\centering
\begin{tabular}{@{}lp{2.1em}p{2.1em}p{2.1em}p{2.1em}p{2.1em}p{2.1em}p{2.1em}p{2.1em}p{2.1em}p{2.1em}@{}}
\toprule
Parameter Budget & \multicolumn{2}{c}{10\%} & \multicolumn{2}{c}{25\%} & \multicolumn{2}{c}{40\%} & \multicolumn{2}{c}{50\%} & \multicolumn{2}{c}{75\%} \\
{\small(Comp.\ Ratio)} & \multicolumn{2}{c}{\small(${\sim}$57\%)} & \multicolumn{2}{c}{\small(${\sim}$45\%)} & \multicolumn{2}{c}{\small(${\sim}$33\%)} & \multicolumn{2}{c}{\small(${\sim}$25\%)} & \multicolumn{2}{c}{\small(${\sim}$6\%)} \\
\cmidrule(l){2-3}\cmidrule(l){4-5}\cmidrule(l){6-7}\cmidrule(l){8-9}\cmidrule(l){10-11}
Method / Model & \textsc{DeiT} & \textsc{Swin} & \textsc{DeiT} & \textsc{Swin} & \textsc{DeiT} & \textsc{Swin} & \textsc{DeiT} & \textsc{Swin} & \textsc{DeiT} & \textsc{Swin}\\
\midrule
\text{Dense} & 15.35 & 3.61 & 65.71 & 60.31 & 74.33 & 80.61 & 79.22 & 83.59 & 81.36 & 85.64\\
\text{Static Sparsity} & 65.26 & 65.6 & 80.06 & 84.37 & 81.48 & \textbf{85.69} & 81.70 & 85.98 & \textbf{81.86} & \textbf{86.23}\\
\text{RigL} & 66.67 & 70.96 & 80.31 & 84.57 & 81.50 & 85.59 & 81.65 & 85.91 & 81.82 & 86.20\\
\text{GMP (\gls{fips})} & \textbf{70.04} & \textbf{74.04} & \textbf{80.64} & \textbf{84.78} & \textbf{81.69} & \textbf{85.69} & \textbf{81.83} & \textbf{85.99} & 81.82 & 86.21\\
\bottomrule
\end{tabular}
\caption{\textbf{Sparsification Method Comparison.} ImageNet top-1 validation accuracy (\%) of \textsc{DeiT-B} (81.85\%)~\citep{DeiT} and \textsc{Swin-L} (86.24\%)~\citep{SWIN} compressed with \gls{fips} using different sparsity methods. GMP consistently outperforms alternatives; the dense baseline collapses at lower budgets.}
\label{tab:app:sparsity_ablation}
\end{table}

\begin{figure}[t]
    \centering
    \begin{subfigure}{0.325\columnwidth}
        \centering
        \includegraphics[width=\textwidth]{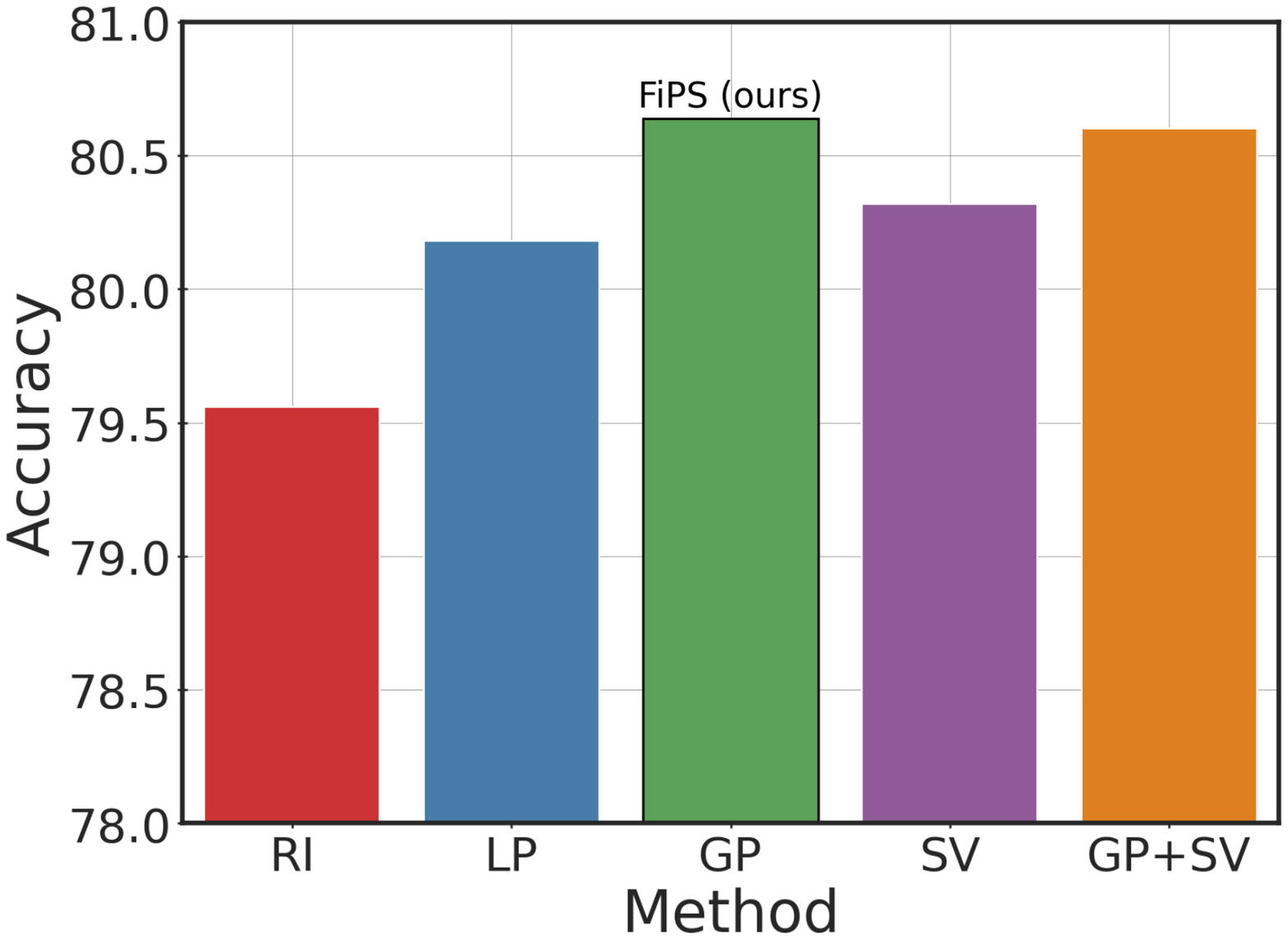}
        \caption{}
        \label{fig:methods_ablation}
    \end{subfigure}
    \hfill
    \begin{subfigure}{0.31\columnwidth}
        \centering
        \includegraphics[width=\textwidth]{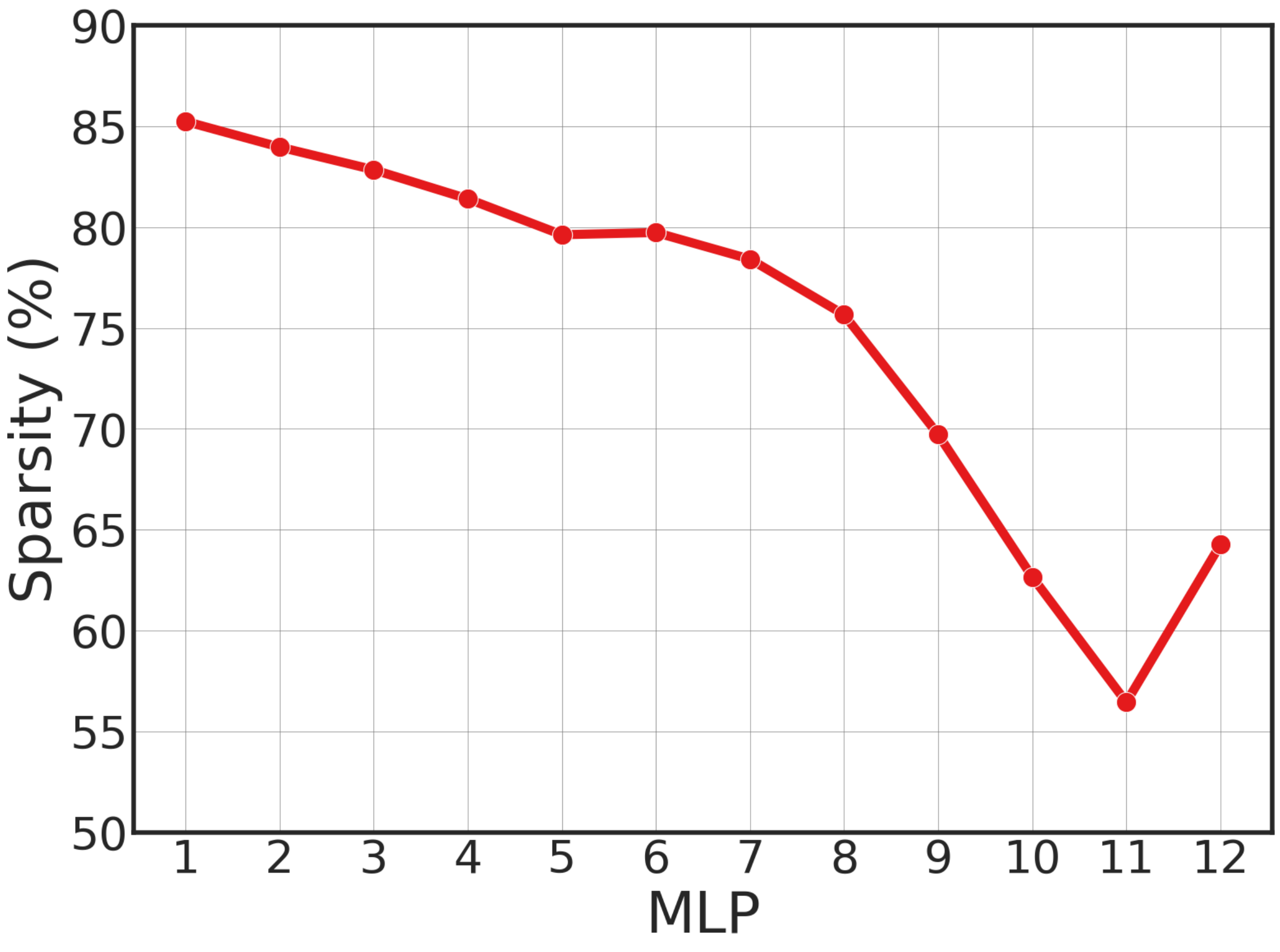}
        \caption{}
        \label{fig:sparsity_dist}
    \end{subfigure}%
    \hfill
    \begin{subfigure}{0.345\columnwidth}
        \centering
        \includegraphics[width=\textwidth]{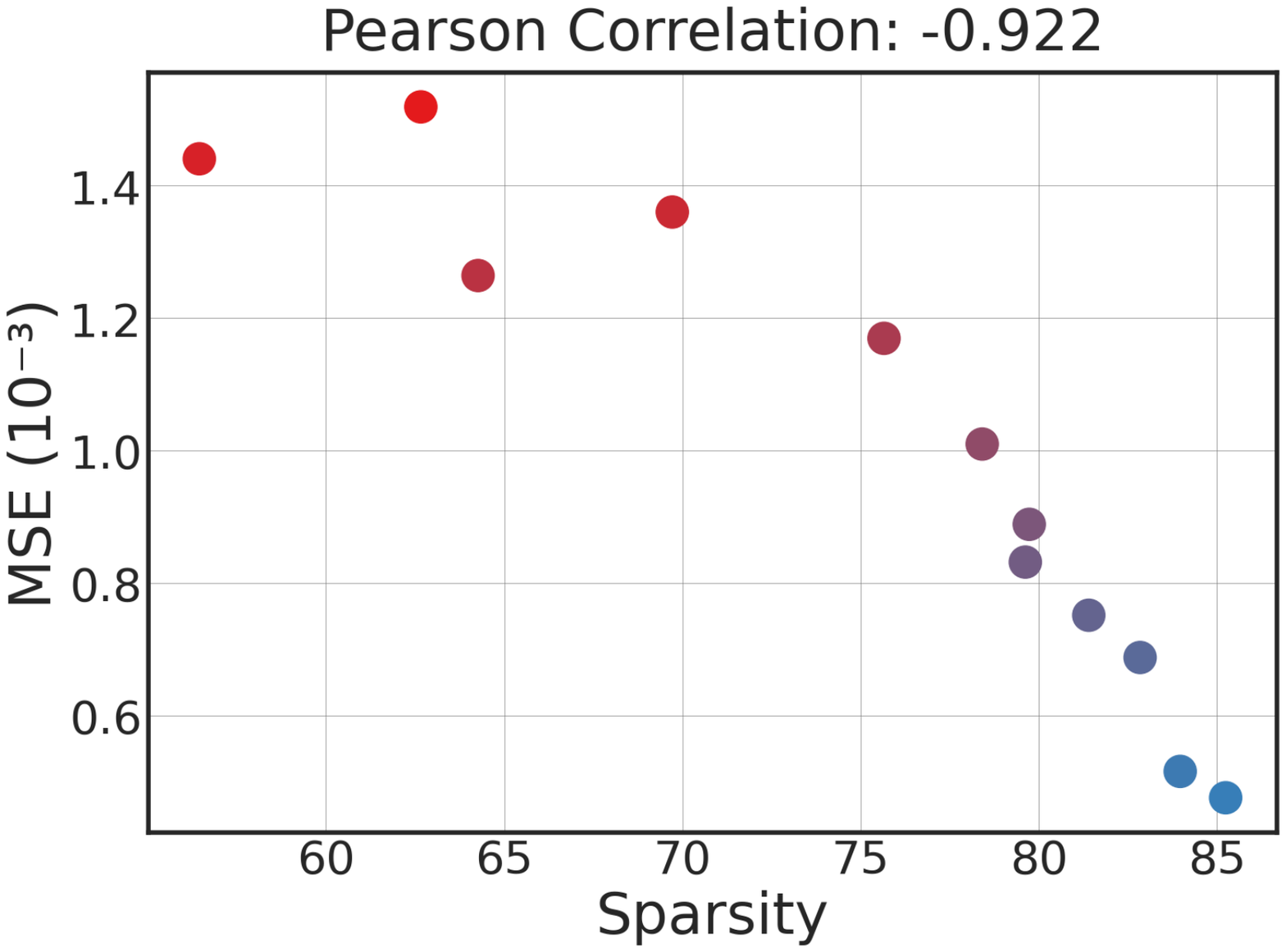}
        \caption{}
        \label{fig:pearson_corr}
    \end{subfigure}
    \caption{\textbf{\textsc{DeiT-B} Ablation and Global Sparsity Analysis.} (\subref{fig:methods_ablation}) Component analysis of the \gls{fips} algorithm: Random Initialization (RI), Local Pruning (LP), Global Pruning (GP), and Scaling Vectors (SV). (\subref{fig:sparsity_dist}) End-of-training sparsity allocation; later layers require more parameters. (\subref{fig:pearson_corr}) Strong correlation between the MSE reported in \Cref{fig:2d_heat} and the parameter distribution captured by \gls{fips}.}
    \label{fig:sparsity_and_corr}
\end{figure}

\paragraph{Sparsity Distribution and MSE Loss.} Initial experiments in~\Cref{fig:2d_heat} reveal that later layers incur higher reconstruction error under uniform compression budgets, suggesting they benefit from greater parameter allocation. \gls{fips} addresses this by applying global magnitude pruning to its sparse factors, and indeed assigns more parameters to later layers (\Cref{fig:sparsity_dist}). Moreover,~\Cref{fig:pearson_corr} shows a strong negative correlation ($-0.922$) between the final sparsity pattern and the MSE losses in~\Cref{fig:2d_heat}, confirming the effectiveness of this adaptive allocation.

\section{Related Work}
\paragraph{Vision Transformers (ViT) \& Large Language Models (LLM).}  
Recent transformer architectures extend beyond the foundational ViT models~\citep{ViT}, which treat image patches as token sequences. \textsc{DeiT}~\citep{DeiT} enhances data efficiency with distillation tokens, while \textsc{Swin}~\citep{SWIN} introduces a hierarchical design using shifted windows. Both vision models employ two FC layers per MLP module. In contrast, decoder-only LLMs such as \textsc{Llama}~\citep{Llama1}, \textsc{Llama-3}~\citep{Llama3}, and \textsc{Gemma-2}~\citep{Gemma2} achieve state-of-the-art zero-shot and instruction-following performance using three FC layers per MLP module in each block.

\paragraph{Sparsity in Neural Networks.}  Early methods involved heuristic pruning, such as removing the smallest-magnitude parameters~\citep{EvaluatingPruningMethods}. Later approaches, such as GMP~\citep{ToPruneOrNotToPrune}, increased the extent of pruning, while dynamic pruning with accelerated schedulers was explored by \citet{howtopruneLM}. Static sparsity uses a preinitialized mask throughout training~\citep{SparsityInDL}, whereas dynamic methods such as RigL~\citep{RigL} adjust the sparsity pattern during training based on gradient information.

\paragraph{Tensor Decomposition.}  
\citet{CompressingTransformers} introduce Adaptive Atomic Feature Mimicking (AAFM) and its global variant Global Feature Mimicking (GFM), which apply truncated PCA on ViT activations followed by fine-tuning. In LLMs, Activation-aware SVD (ASVD)~\citep{ASVD}, truncation-aware SVD-LLM~\citep{SVDLLM}, and SVD-LLM V2 with rank distillation and layer-wise allocation~\citep{SVDLLMv2} serve as our baselines.

\paragraph{Parameter Sharing.}
Several prior works explore parameter sharing in neural networks. \citet{StructuredMultiHashing} introduce a Sum-Product reducer to map shared parameters, and \citet{TBasis} employ TR decomposition for shared parameters in 3D tensors. At the block level, ALBERT~\citep{ALBERT} shares entire transformer blocks (all parameters across layers), reducing model size but constraining every layer to identical weights. \citet{MiniViT} propose ``Weight Multiplexing,'' sharing parameters between MLP modules in ViTs with distillation and learned linear projections between blocks to aid recovery; however, MiniViT does not impose sparsity on its projection factors and relies on distillation for quality. Concurrent work by \citet{BasisSharing} also explores cross-layer parameter sharing via SVD, representing weight matrices as linear combinations of shared basis vectors with dense, layer-specific coefficients.

In summary, \gls{fips} is distinguished from all the above compression approaches by combining three properties that no prior method jointly offers: (i)~cross-block weight sharing via a shared basis $\mathbf{U}$, (ii)~low-rank factorization initialized by SVD, and (iii)~sparsity on the projection matrices $\mathbf{V}_i$, optimized via block-wise reconstruction loss. In particular, we show in \Cref{sec:OptimalDecomp} that enforcing sparsity in $\mathbf{V}$ is critical for achieving lower reconstruction error at the same parameter budget compared to dense coefficients~\citep{BasisSharing}. Methods such as pruning combined with distillation~\citep{howtopruneLM} and LoRA-based compression~\citep{hu2021loralowrankadaptationlarge} operate in a complementary regime---they rely on auxiliary supervision signals or learned adapters---and are orthogonal to \gls{fips}, as we demonstrate with QAT in \Cref{tab:qat_table}.

\section{Conclusion}
We presented \gls{fips}, a framework for compressing transformer MLPs via fine-grained inter-layer parameter sharing that unifies cross-block weight tying, low-rank factorization, and sparsity in a single optimization. \gls{fips} achieves state-of-the-art compression--accuracy trade-offs on MLP layers: up to 33\% compression on ViTs with $<$1\% accuracy loss (up to 57\% with fine-tuning), up to 20\% on LLMs while outperforming existing SVD-based methods at matched compression, and 8$\times$ on \textsc{Gemma-2-2B} when combined with QAT. These findings establish parameter sharing as a competitive alternative to existing compression strategies. Future work includes extending to attention layers---whose projection matrices share the same block-repeated structure that \gls{fips} exploits---quantizing the shared bases, and developing specialized kernels that keep $\mathbf{U}$ resident in fast memory for efficient on-device inference.

\section*{Broader Impact Statement}
This paper advances Machine Learning by introducing Fine-grained Parameter Sharing (\gls{fips}), a model compression method that improves the efficiency of Vision Transformers (ViTs) and Large Language Models (LLMs). By leveraging parameter sharing, low-rank factorization, and sparsity, \gls{fips} reduces computational and memory costs, enhancing AI accessibility on resource-constrained devices. While model compression promotes efficiency and sustainability, it may also enable broader AI deployment in sensitive domains with ethical concerns such as bias, misinformation, and privacy. Nevertheless, this work should not introduce new risks beyond those inherent in deep learning, but we encourage responsible deployment and ethical considerations in practice.

\section*{Author Contributions}
Cem Üyük led the project, proposed and executed the experimental plan, facilitated the team meetings, developed the software architecture, implemented static sparse training and provided code review for the sparse training algorithms, wrote the initial draft of the paper, and further contributed to writing significantly while also creating most of the plots. Mike Lasby implemented sparse training algorithms, assisted the software architecture development, handled distributed training integration, performed code reviews, and assisted with writing and proofreading the paper. Mohamed Yassin assisted with coding and running inference experiments. Utku Evci proposed the project and its central idea, contributed to the research plan and direction, advised Cem, reviewed the code, helped substantially with the writing, and created some of the plots. Yani Ioannou helped with the research direction, contributed to the paper’s motivation, helped with the writing, provided compute resources, and supervised the work by members of the Calgary ML Lab at the University of Calgary, including Cem Üyük (Visiting Student Researcher), Mike Lasby (PhD Student), and Mohamed Yassin (Research Assistant).

\section*{Acknowledgments}
We gratefully acknowledge the support of Alberta Innovates (ALLRP 577350-22, ALLRP 600038-24), the Natural Sciences and Engineering Research Council of Canada (NSERC) (RGPIN-2022-03120, DGECR-2022-00358), Defence Research and Development Canada (DGDND-2022-03120), and NSERC/Agence Nationale de la Recherche (ANR) (ALLRP 602719-24). This project was undertaken thanks to funding from IVADO and the Canada First Research Excellence Fund. This research was enabled in part by support provided by the Digital Research Alliance of Canada (alliancecan.ca) and Google Cloud. We also acknowledge Erik Schultheis' very helpful feedback with regard to custom kernel design.

\bibliography{main}
\bibliographystyle{tmlr}

\appendix
\section{Appendix}
\subsection{Method}\label{app:Concat}

Referring to~\Cref{sec:WhichDimsToShare}, we detail four methods for concatenating the FC weights of two MLPs (four matrices \(\mathbf{W}_{ij}\in\mathbb{R}^{d\times p}\), with \(p=4d\) and \(i,j\in\{1,2\}\)):

\begin{enumerate}[label=\Roman*.]
    \item \textbf{Full long‐axis concatenation}, forming
    \[
      \mathbf{W}
      = [\,\mathbf{W}_{11}\,\mathbf{W}_{12}\,\mathbf{W}_{21}\,\mathbf{W}_{22}\,]
      \;\in\;\mathbb{R}^{d\times 16d}.
    \]
    
    \item \textbf{Module‐wise long + inter‐module short:}
    \newline
    \[\mathbf{W}_A=[\mathbf{W}_{11}\mathbf{W}_{12}]\in\mathbb{R}^{d\times 8d}\] 
    \newline
    \[\mathbf{W}_B=[\mathbf{W}_{21}\mathbf{W}_{22}]\in\mathbb{R}^{d\times 8d}\]
    \newline
    \[
      \mathbf{W}
      = \begin{bmatrix}\mathbf{W}_A \\[-2pt] \mathbf{W}_B\end{bmatrix}
      \;\in\;\mathbb{R}^{2d\times 8d}.
    \]
    
    \item \textbf{Module‐wise short + inter‐module long:}
    \newline
    \[\mathbf{W}_C=[\mathbf{W}_{11}\mathbf{W}_{21}]\in\mathbb{R}^{2d\times 4d}\]
    \newline
    \[\mathbf{W}_D=[\mathbf{W}_{12}\mathbf{W}_{22}]\in\mathbb{R}^{2d\times 4d}\],
    \newline
    \[
      \mathbf{W}
      = [\,\mathbf{W}_C\,\mathbf{W}_D\,]
      \;\in\;\mathbb{R}^{2d\times 8d}.
    \]
    
    \item \textbf{Full short‐axis concatenation}, yielding
    \[
      \mathbf{W}
      = \begin{bmatrix}
        \mathbf{W}_{11} \\[-2pt] \mathbf{W}_{12} \\[-2pt] \mathbf{W}_{21} \\[-2pt] \mathbf{W}_{22}
      \end{bmatrix}
      \;\in\;\mathbb{R}^{4d\times 4d}.
    \]
\end{enumerate}

These configurations are evaluated in~\Cref{fig:concat_dims} and discussed in~\Cref{sec:WhichDimsToShare}.

\begin{figure*}[tbp]
    \centering
    \includegraphics[width=0.9\textwidth]{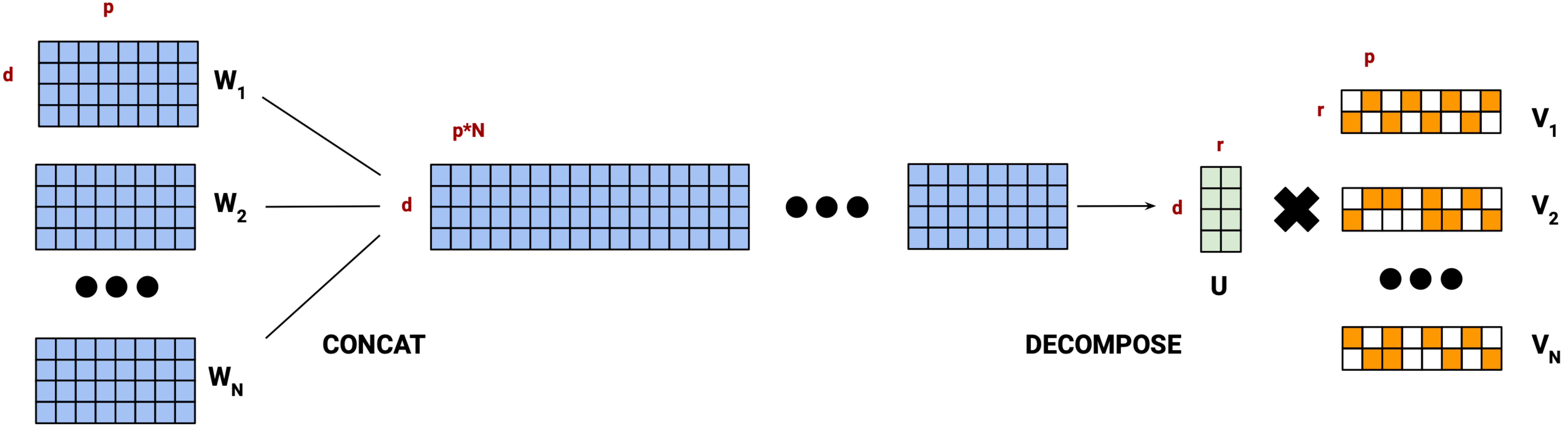} 
    \caption{\textbf{Parameter Sharing Through Sparse Tensor Decomposition.} A group of FC layers are concatenated along the larger dimension, $p$, and decomposed into two matrices: a shared basis, $\textbf{U}$, and a sparse projection matrix, which is then sliced up respectively for each layer.}
    \label{fig:CompressionInit}
\end{figure*}
The overall method is depicted in~\Cref{fig:CompressionInit}, and further details are explained below.

\subsubsection{Growing Neurons in Shared Bases and Sparse Factors}\label{app:growNeurons}
As discussed in~\Cref{sec:method}, high parameter budgets and sparsity levels (e.g., 26.5\% parameter budget, 75\% sparsity, and groups of four blocks in \textsc{DeiT-B}) often result in the rank $r$ exceeding the model dimension $d$. Since SVD yields only $d$ initialization directions, we investigate three methods to initialize the remaining $k = r - d$ dimensions:
\begin{enumerate}
    \item \textbf{Random Growth:} Initialize new neurons in $\mathbf{U}$ to zero and in $\mathbf{V}$ randomly using \citet{KaimingHeInit};
    \item \textbf{Neuron Splitting:} Duplicate the top $k$ neurons of $\mathbf{U}$ and halve the top $k$ neurons of $\mathbf{V}$, following \citet{Net2Net};
    \item \textbf{Hybrid Initialization:} Initialize new neurons in $\mathbf{U}$ to zero and derive those in $\mathbf{V}$ from the top $k$ neurons, scaled by $\tau$. This minimizes the immediate impact of new neurons in $\mathbf{V}$, allowing their gradual reactivation, as proposed by \citet{GradMax}.
\end{enumerate}
After performing a hyperparameter sweep for $\tau$, hybrid initialization outperformed the alternatives, achieving 1\% and 2\% higher accuracy than methods (1) and (2), respectively.

\subsection{Compression Ratio Computation}
\label{app:compression_ratio}

The reported compression ratio accounts for all storage overhead, including sparse metadata and uncompressed components:

\begin{itemize}
    \item \textbf{Sparse factors (\(\mathbf{V}\)):} We store both the nonzero values and their positions. For unstructured sparsity, we use a bitmap mask (1 bit per element). For 2:4 structured sparsity, the fixed pattern eliminates the need for explicit indices.
    \item \textbf{Shared basis (\(\mathbf{U}\)):} Each group has one shared basis \(\mathbf{U}\), counted once per group. For example, with \(\beta = [4, 6, 6, 6, 6, 4]\) for \textsc{Llama-7B} (6 groups), we store 6 shared bases.
    \item \textbf{Non-MLP parameters:} Attention projections, embeddings, and layer norms remain uncompressed and are counted at full precision.
    \item \textbf{Quantized setup:} Each parameter is counted at its precision level (e.g., 3 bits for INT3), and masks are counted as 1 bit per element.
\end{itemize}

Formally, the compression ratio is:
\begin{equation}
    \text{Compression Ratio} = \frac{|\theta_{\text{non-MLP}}| \cdot b + G \cdot |\mathbf{U}| \cdot b + \sum_{i} \left[(1-s) \cdot |\mathbf{V}_i| \cdot b + |\mathbf{V}_i| \cdot 1\right]}{|\theta_{\text{original}}| \cdot b_{\text{original}}},
\end{equation}
where \(G\) is the number of groups, \(s\) is the sparsity level, \(b\) is the bits per parameter (e.g., 16 for \texttt{bfloat16}), \(b_{\text{original}}\) is the original precision, and the sum is over all sparse factors \(\mathbf{V}_i\) across all layers. For 2:4 structured sparsity, the mask term \(|\mathbf{V}_i| \cdot 1\) is omitted as the sparsity pattern is implicit.

\subsection{Model Links}
\label{app:models}
\begin{itemize}
    \item \textsc{DeiT-B}~\citep{DeiT}: \url{https://huggingface.co/facebook/deit-base-patch16-224}
    \item \textsc{Swin-L}~\citep{SWIN}: \url{https://huggingface.co/microsoft/swin-large-patch4-window7-224}
    \item \textsc{Gemma-2-2B-IT}~\citep{Gemma2}: \url{https://huggingface.co/google/gemma-2-2b-it}
    \item \textsc{Gemma-2-9B}~\citep{Gemma2}: \url{https://huggingface.co/google/gemma-2-9b}
    \item \textsc{Llama-7B}~\citep{Llama1}: \url{https://huggingface.co/huggyllama/llama-7b}
    \item \textsc{Llama-3.1-8B}~\citep{Llama3}: \url{https://huggingface.co/meta-Llama/Llama-3.1-8B}
\end{itemize}

\subsection{Further LLM Results}
\begin{table}[H]
\centering
\begin{tabular}{lcc}
\toprule
\textbf{Variant} & \textbf{Comp.\ Ratio} & \textbf{PPL} $\downarrow$ \\
\midrule
Original  & 0\% & 7.34 \\
\midrule
SVD-LLM  & 20\% & 15.84 \\
SVD-LLM V2  & 20\% & 11.72 \\
\gls{fips} (ours) & 20\% & \textbf{8.10} \\
\midrule
SVD-LLM  & 40\% & 75.42 \\
\gls{fips} (ours) & 40\% & \textbf{10.57} \\
\bottomrule
\end{tabular}
\caption{Perplexity on C4 for Llama7B under different compression ratios applied with \gls{fips} and the baselines.}
\label{app:tab:llama7b_wikitext2}
\end{table}

\subsection{Different Sparsification Methods}
\label{app:sparsification_methods}
The sparsification method comparison is presented in \Cref{tab:app:sparsity_ablation} in the main text (\S\ref{Sec:Ablation}). For \textsc{DeiT-B}, \textit{RigL} consistently outperforms both \textit{Dense} and \textit{Static Sparsity} across parameter budgets ranging from 10\% to 50\%. At higher parameter budgets, all methods converge to similar accuracies approaching the original model's performance. For \textsc{Swin-L}, \textit{RigL} surpasses \textit{Dense} and \textit{Static Sparsity} at 10\% and 25\% parameter budgets. However, at higher parameter budgets, \textit{Static Sparsity} achieves slightly higher accuracies.

\subsection{Structured Sparsity}
\label{app:structured_sparsity}
We evaluate the generalization performance of \gls{fips} using structured sparsity, with results presented in~\Cref{tab:app:structured_sparsity}. The methods evaluated include the Straight Through Estimator (\textit{STE}), which employs top-$k$ weight magnitude selection, projects parameters into a sparse subspace during training, and applies gradients to dense parameters through a gradual pruning schedule; Sparse-Refined STE (\textit{SR-STE})~\citep{SRSTE}, which mitigates the adverse effects of approximated gradients; and N:M Structured RigL (\textit{NMSRigL}) and N:M Structured GMP (\textit{NMSGMP})~\citep{JaxPruner, SRigL}, where N:M specifies the sparsity pattern of the weight matrix (e.g., 50\% sparsity in FC matrices of size $d \times 4d$ corresponds to a 2:4 structure).
\begin{table}[tbp]
\centering
\small
\setlength{\tabcolsep}{5pt}
\renewcommand{\arraystretch}{1.1}
\[
\begin{tabular}{@{}lccccc@{}}
\toprule
\textbf{Comp.\ Ratio} & \textbf{10\%} & \textbf{25\%} & \textbf{40\%} & \textbf{50\%} & \textbf{75\%} \\
\midrule
STE                   & 42.89 & 73.26 & 78.26 & 79.36 & 78.89 \\
SR-STE                & 45.31 & 75.53 & 79.71 & 80.68 & 81.24 \\
NMSRigL               & 44.87 & 75.71 & 79.97 & 80.99 & 81.40 \\
NMSGMP                & 52.36 & 76.88 & 80.59 & 81.31 & 81.51 \\
\gls{fips} (50\% Sparsity)  & 54.00 & 77.56 & 80.94 & 81.63 & 81.77 \\
\midrule
\gls{fips} (75\% Sparsity)  & 70.04 & 80.64 & 81.69 & 81.83 & 81.82 \\
\bottomrule
\end{tabular}
\]
\caption{\textbf{Structured Sparsity Performance.} ImageNet top-1 accuracy (\%) of \textsc{DeiT-B} (81.85\%)~\citep{DeiT} for various structured sparsification methods at 50\% and 75\% sparsity, compared to unstructured \gls{fips}. Methods include Straight Through Estimator (\textit{STE}), Sparse-Refined STE, N:M Structured RigL (\textit{NMSRigL}), and N:M Structured GMP (\textit{NMSGMP}) at 50\% sparsity, corresponding to 2:4 structures~\citep{JaxPruner, SRSTE, SRigL}.}
\label{tab:app:structured_sparsity}
\end{table}

\subsection{Sensitivity Analysis}
\label{app:sensitivity}
The sensitivity analysis plots are presented in \Cref{fig:sensitivity} in the main text (\S\ref{Sec:Ablation}).

\paragraph{Calibration Dataset Size and Training Length.}
We examine how the number of calibration batches and training epochs affects performance using a fixed batch size of 128. To ensure at least one example from each category, we begin with a minimum of 10 batches and also evaluate 20, 40, and 80 batches. After filtering out configurations more than 0.25\% below the highest accuracy, we adopt the most efficient setting of 20 epochs over 20 batches for all reported results, as shown in~\Cref{fig:epoch_batch_sweep}.

\paragraph{Optimal Sparsity for Sparse Factors.} We compressed \textsc{DeiT-B} as described in~\Cref{sec:ViTExperiments}, using sparsity levels ranging from 50\% to 96.9\% (\Cref{fig:sparsity_sweep}). The best performance was observed at 75\% sparsity. While increasing sparsity to 87\% yielded similar accuracy, lowering it to 50\% resulted in a notable performance drop, likely due to a significant reduction in rank.

\subsection{Hyper-parameters}\label{app:hyperParams}
\subsubsection{Ablation on the Block–Grouping Hyper-parameter $\beta$}
\label{app:block_groups} 

\paragraph{Definition.}
We define $\beta$ as an \emph{ordered list} whose \(i^{\text{th}}\) element gives the number of consecutive decoder blocks whose MLP weights are tied in the \(i^{\text{th}}\) parameter sharing group:
\[
\beta=\bigl[\beta_1,\;\beta_2,\;\dots,\;\beta_G\bigr],\quad\textstyle\sum_{g=1}^{G}\beta_g=L,
\]
where \(L\) is the total number of decoder blocks.  
Self-attention parameters remain untied in all experiments.  
For every architecture we sweep over a small number of plausible \(\beta\) lists (3–5 candidates) and keep the one with the lowest validation perplexity (PPL) after compression.

\paragraph{Default Heuristic.}
Our analysis in~\Cref{sec:LayerSharing} shows that adjacent layers exhibit smaller reconstruction error when sharing, and deeper layers require more capacity. This motivates a ``tapered'' grouping where group sizes are smaller at the network extremes and larger in the middle. Based on our experiments, we propose the following reproducible default for a decoder with \(L\) blocks:
\begin{quote}
Use 4–6 groups with \(\beta\) tapered as \([\textit{small}, \textit{medium}, \dots, \textit{medium}, \textit{small}]\), where \(\textit{small} \approx L/8\) and \(\textit{medium} \approx L/5\).
\end{quote}
For example:
\begin{itemize}
    \item \(L=32\) (Llama): \(\beta = [4, 6, 6, 6, 6, 4]\)
    \item \(L=26\) (Gemma-2-2B): \(\beta = [4, 4, 5, 5, 4, 4]\)
    \item \(L=12\) (DeiT-B): \(\beta = [4, 4, 4]\)
\end{itemize}

\paragraph{Block Groups of ViTs.}
Using the list-valued notation for $\beta$ introduced above, we set
\[
\beta_{\textsc{DeiT-B}}=[4,\,4,\,4],
\]
i.e.\ three groups of four consecutive blocks (each block contains one MLP).

The depth pattern of \textsc{Swin-L} is \(2{+}2{+}18{+}2\) blocks.
We tie MLP weights inside every 2-block stage
and split the 18-block stage into three groups of six,
which gives
\[
\beta_{\textsc{Swin-L}}=[2,\,2,\,6,\,6,\,6,\,2].
\]

\paragraph{Gemma-2-2B.}  
Table~\ref{tab:gemma-groups} shows the five \(\beta\) lists evaluated for \textsc{Gemma-2-2B-it} at 20\% compression.  
Config.~1—\(\beta=[\,4,4,5,5,4,4\,]\)—yields the lowest PPL and is therefore used in the main paper. Notably, alternative groupings yield PPL within $\sim$2 points of the optimal (21.42 vs.\ $\approx$23), indicating that \gls{fips} is not overly sensitive to this hyperparameter.

\begin{table}[!t]
\centering
\caption{\textbf{Block–grouping sweep for \textsc{Gemma-2-2B}.}  
Each row lists the candidate \(\beta\) and the resulting validation perplexity (PPL) on WikiText-2.  
Baseline (no sharing) PPL is \(15.61\); lower is better.}
\label{tab:gemma-groups}
\begin{tabular}{@{}lc@{}}
\toprule
\textbf{Config.} & \textbf{\(\beta\) list} \hspace{1em} PPL $\downarrow$\\
\midrule
1 & $[\,4,4,5,5,4,4\,]$ \hfill 21.42\\
2 & $[\,1,4,4,4,4,4,4,1\,]$ \hfill $\approx$23\\
3 & $[\,2,3,4,4,4,4,3,2\,]$ \hfill $\approx$22\\
4 & $[\,2,6,6,6,6\,]$ \hfill 21.81\\
5 & $[\,3,5,5,5,5,3\,]$ \hfill 21.86\\
\bottomrule
\end{tabular}
\end{table}

\paragraph{Llama-2-7B.}  
With \(L=32\) decoder blocks we compared three \(\beta\) lists:

\begin{center}
\setlength{\tabcolsep}{6pt}
\renewcommand{\arraystretch}{1.1}
\begin{tabularx}{\linewidth}{@{}l X l@{}}
\toprule
\textbf{Candidate \(\beta\)} & \textbf{Block groups (sizes)} & \textbf{Observation}\\
\midrule
$[\,4,4,4,4,4,4,4,4\,]$ & 8 groups × 4 blocks & Highest PPL\\[2pt]
$\boldsymbol{[\,4,6,6,6,6,4\,]}$ & \textbf{6 groups with sizes} $(4,6,6,6,6,4)$ & \textbf{Best PPL; used in \S\ref{Sec:RecoveryResults}}\\[2pt]
$[\,8,8,8,8\,]$ & 4 groups × 8 blocks & Slightly worse than above\\
\bottomrule
\end{tabularx}
\end{center}

A gently varying list—with smaller groups at the extremes and larger groups in the middle—provides the best compression/accuracy trade-off.

\paragraph{Llama-3.1-8B.}  
The 8B model shares the same 32-layer decoder.  
We reused the winning \(\beta\) from the 7B sweep,
\[
\beta_{\mathrm{opt}}=[\,4,6,6,6,6,4\,],
\]
because (i) it keeps the total tied-parameter ratio identical and (ii) in a spot-check it preserved PPL within +3.5 of the uncompressed baseline—better than the uniform alternatives \([\,8,8,8,8\,]\) or \([\,4,4,4,4,4,4,4,4\,]\). This demonstrates that the optimal \(\beta\) transfers across models of similar architecture.

\paragraph{Key Insights.}
\begin{itemize}
\item Optimal \(\beta\) often starts and ends with smaller groups, reflecting the intuition that early and late layers contain more specialized features.
\item Extremely fine-grained sharing (e.g., many 4-block groups) hurts accuracy, while overly coarse sharing (uniform 8-block groups) sacrifices capacity.
\item For Llama models, a tapered list such as \([4,6,6,6,6,4]\) ties roughly 22–25\% of MLP parameters yet adds only \(\sim\)3–4 perplexity points.
\item The method is robust to the choice of \(\beta\): alternative groupings typically yield performance within 2 PPL points of the optimal.
\end{itemize}

These ablations inform all main-text compression results.
\subsubsection{Optimizer}
\paragraph{ViT Compression.} To minimize local error, we employ a logarithmic grid for hyper-parameter tuning. The learning rates for Dense, Static Sparsity, GMP, and RigL are set as follows for both \textsc{DeiT-B} and \textsc{Swin-L}:
\begin{enumerate}
    \item Dense: $1.25 \times 10^{-4}$,
    \item Static Sparsity: $2.5 \times 10^{-4}$,
    \item GMP: $1 \times 10^{-3}$,
    \item RigL: $1 \times 10^{-3}$.
\end{enumerate}

\paragraph{ViT Transfer Learning.} We use a linear grid, as some hyper-parameters are derived from the codebase of \textsc{DeiT}. The optimal learning rates for \gls{fips} are:
\begin{enumerate}
    \item CIFAR-100: $2.5 \times 10^{-5}$;
    \item Flowers102: $1 \times 10^{-4}$;
    \item Oxford-III-Pets: $7.5 \times 10^{-6}$;
    \item iNaturalist 2019: $1 \times 10^{-4}$.
\end{enumerate}

\paragraph{LLMs.} Eight logarithmically spaced values were swept. The final values for \gls{fips} are:
\begin{enumerate}
    \item \textsc{Gemma-2-2B}: $4.0 \times 10^{-4}$,
    \item \textsc{Llama-7B}: $1.0 \times 10^{-6}$,
    \item \textsc{Llama-3.1-8B}: $1.0 \times 10^{-5}$.
\end{enumerate}

\paragraph{Sparsifier}
\subparagraph{Gradual Magnitude Pruning (GMP).} 
GMP begins with an initial sparsity level of 25\%. During training, sparsity is gradually increased to 50\% at the 25\% training mark and ultimately reaches 75\% by training completion. We use $\Delta T = 50$ for update steps.

\subparagraph{RigL.}
RigL employs an initialization phase that combines pruning with a growth ratio of $0.1$ for block-wise error minimization and a growth ratio of $0.05$ for transfer learning tasks, with $\Delta T = 50$ for growth and pruning steps. This conservative growth ratio in transfer learning helps preserve the mask obtained during initial training, ensuring that learned masks are retained.

\begin{figure*}[tbp]
  \centering
  \begin{subfigure}[t]{0.48\textwidth}
    \centering
    \includegraphics[width=\linewidth]{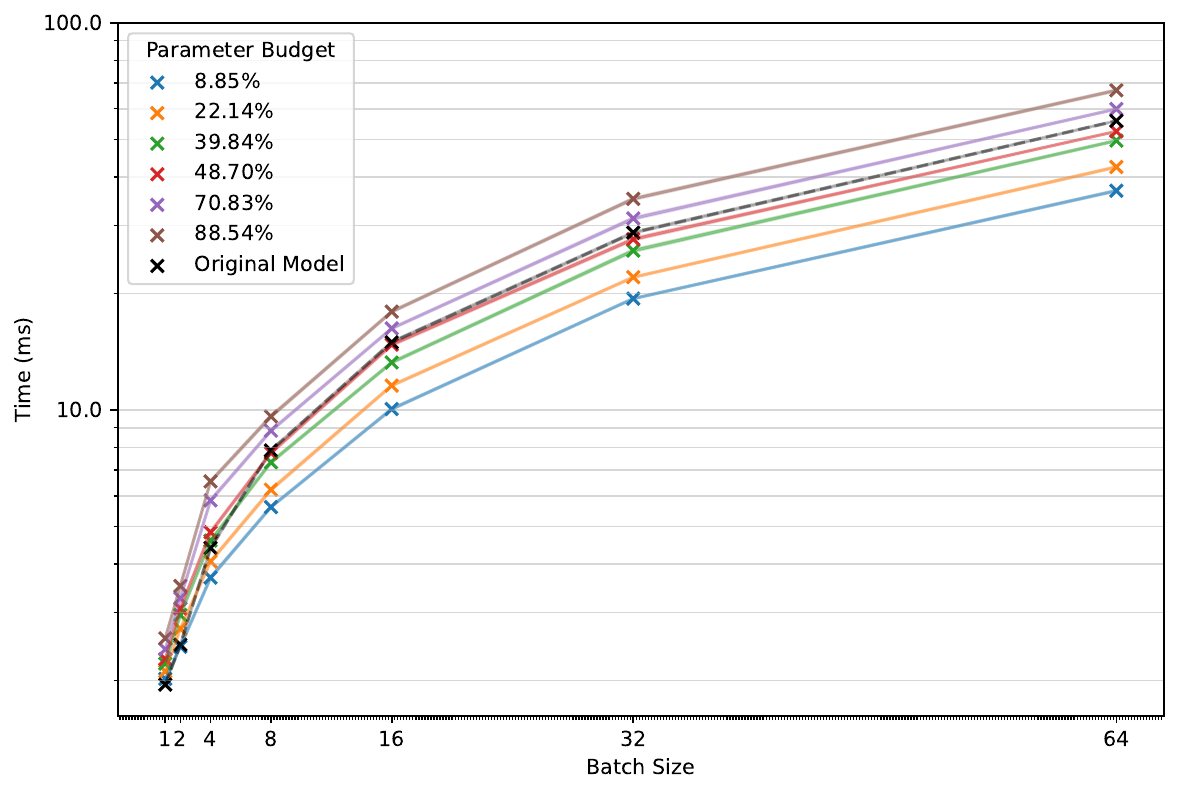}
    \caption{GPU latency (ms) vs.\ parameter budget (\%).}
    \label{fig:two-four}
  \end{subfigure}
  \hfill
  \begin{subfigure}[t]{0.48\textwidth}
    \centering
    \includegraphics[width=\linewidth]{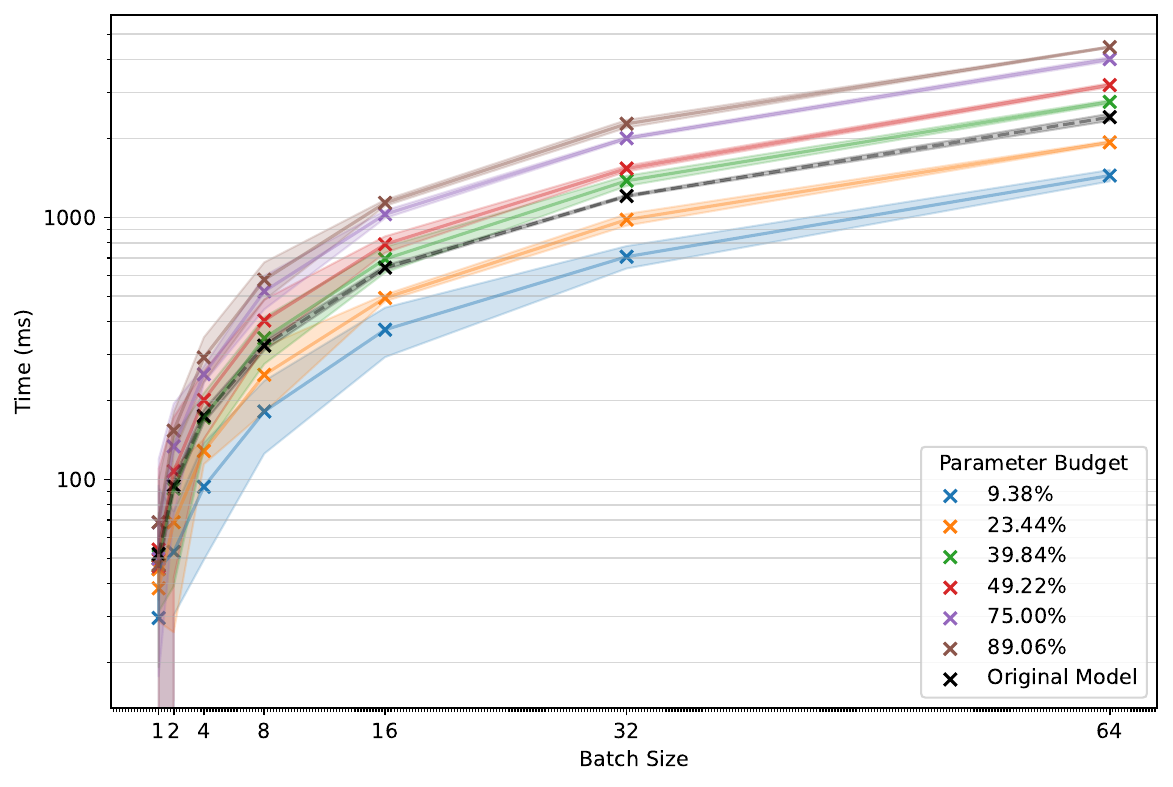}
    \caption{CPU latency (ms) vs.\ parameter budget (\%).}
    \label{fig:deepsparse}
  \end{subfigure}

  \caption{\textbf{\textsc{DeiT-B} inference latency benchmarks.}
  (a) End-to-end latency of 2:4 sparse \gls{fips} on an NVIDIA A4000 for batch sizes 1–64; a 22\,\% parameter budget yields a 25\,\% speed-up once the batch size exceeds 8.
  (b) Latency of 75\,\% unstructured sparse \gls{fips} accelerated by DeepSparse on an Intel Xeon W-2145 CPU, outperforming the dense model at every tested batch size. Shaded regions denote $\pm 1\sigma$ over runs collected by \texttt{torch.benchmark.Timer.blocked\_autorange}; GPU timings exclude \texttt{torch.compile} warmup iterations.}
  \label{fig:latency-benchmarks}
\end{figure*}

\begin{figure*}[!h]
    \centering
    \includegraphics[width=0.9\linewidth]{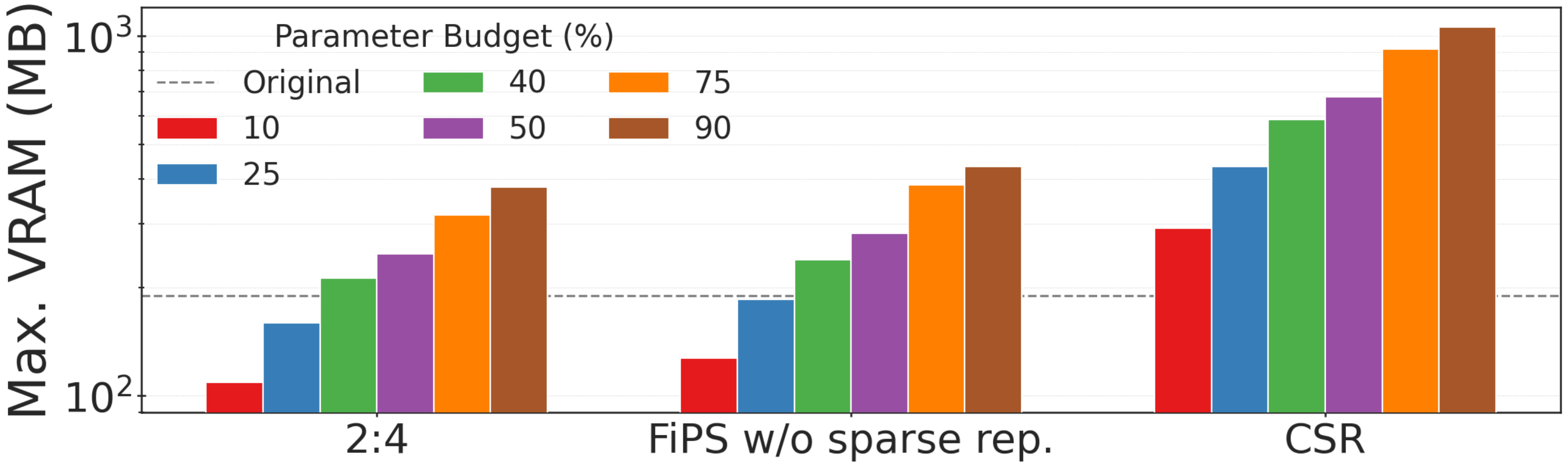}
    \caption{\textbf{\textsc{DeiT-B} Inference Memory Profile.}  
    Maximum VRAM allocation at batch size 64 for 50\% sparse \gls{fips} using 2:4, strided (dense), and CSR tensor formats. At 10\% and 25\% parameter budgets, 2:4 sparsity reduces peak memory by 44\% and 18\%, respectively; CSR incurs higher overhead at modest sparsity due to index storage.}
    \label{fig:memory-profile}
\end{figure*}

\subsection{Latency and Memory Profiling}
As discussed in~\Cref{par:shared-init}, high levels of sparsity and parameter budgets can result in the SVD rank exceeding a model's hidden dimension. For instance, in \textsc{DeiT-B}, achieving 75\% sparsity under parameter budget constraints exceeding 26.5\% with four block groups increases the rank of the shared singular vectors beyond the original model's embedding dimension. Efficient sparse operations and representations are crucial for minimizing the latency and memory overhead introduced by \gls{fips}.~\Cref{fig:latency-benchmarks} and~\Cref{fig:memory-profile} summarize the latency and memory results for \textsc{DeiT-B} compressed with \gls{fips} using $2{:}4$ structured GMP, highlighting the resulting speedups and memory savings on both CPU and GPU platforms.
\label{app:latency}

\end{document}

%% file: math_commands.tex
\usepackage{amsmath,amsfonts,bm}

\def\eqref#1{equation~\ref{#1}}

\def\1{\bm{1}}

\DeclareMathAlphabet{\mathsfit}{\encodingdefault}{\sfdefault}{m}{sl}
\SetMathAlphabet{\mathsfit}{bold}{\encodingdefault}{\sfdefault}{bx}{n}



%% file: main.bib
@misc{DeiT,
      title={Training data-efficient image transformers \& distillation through attention}, 
      author={Hugo Touvron and Matthieu Cord and Matthijs Douze and Francisco Massa and Alexandre Sablayrolles and Hervé Jégou},
      year={2021},
      eprint={2012.12877},
      archivePrefix={arXiv},
      primaryClass={cs.CV},
      url={https://arxiv.org/abs/2012.12877}, 
}

@misc{SWIN,
      title={Swin Transformer: Hierarchical Vision Transformer using Shifted Windows}, 
      author={Ze Liu and Yutong Lin and Yue Cao and Han Hu and Yixuan Wei and Zheng Zhang and Stephen Lin and Baining Guo},
      year={2021},
      eprint={2103.14030},
      archivePrefix={arXiv},
      primaryClass={cs.CV},
      url={https://arxiv.org/abs/2103.14030}, 
}

@misc{ViT,
      title={An Image is Worth 16x16 Words: Transformers for Image Recognition at Scale}, 
      author={Alexey Dosovitskiy and Lucas Beyer and Alexander Kolesnikov and Dirk Weissenborn and Xiaohua Zhai and Thomas Unterthiner and Mostafa Dehghani and Matthias Minderer and Georg Heigold and Sylvain Gelly and Jakob Uszkoreit and Neil Houlsby},
      year={2021},
      eprint={2010.11929},
      archivePrefix={arXiv},
      primaryClass={cs.CV},
      url={https://arxiv.org/abs/2010.11929}, 
}

@inproceedings{CompressingTransformers,
author = {Yu, Hao and Wu, Jianxin},
title = {CompressingTransformers},
year = {2023},
isbn = {978-1-57735-880-0},
publisher = {AAAI Press},
url = {https://doi.org/10.1609/aaai.v37i9.26304},
doi = {10.1609/aaai.v37i9.26304},
booktitle = {Proceedings of the Thirty-Seventh AAAI Conference on Artificial Intelligence and Thirty-Fifth Conference on Innovative Applications of Artificial Intelligence and Thirteenth Symposium on Educational Advances in Artificial Intelligence},
articleno = {1235},
numpages = {9},
series = {AAAI'23/IAAI'23/EAAI'23}
}

@misc{SparsityInDL,
      title={Sparsity in Deep Learning: Pruning and growth for efficient inference and training in neural networks}, 
      author={Torsten Hoefler and Dan Alistarh and Tal Ben-Nun and Nikoli Dryden and Alexandra Peste},
      year={2021},
      eprint={2102.00554},
      archivePrefix={arXiv},
      primaryClass={cs.LG},
      url={https://arxiv.org/abs/2102.00554}, 
}

@misc{ToPruneOrNotToPrune,
      title={To prune, or not to prune: exploring the efficacy of pruning for model compression}, 
      author={Michael Zhu and Suyog Gupta},
      year={2017},
      eprint={1710.01878},
      archivePrefix={arXiv},
      primaryClass={stat.ML},
      url={https://arxiv.org/abs/1710.01878}, 
}

@misc{howtopruneLM,
      title={How to Prune Your Language Model: Recovering Accuracy on the "Sparsity May Cry'' Benchmark}, 
      author={Eldar Kurtic and Torsten Hoefler and Dan Alistarh},
      year={2023},
      eprint={2312.13547},
      archivePrefix={arXiv},
      primaryClass={cs.CL},
      url={https://arxiv.org/abs/2312.13547}, 
}

@misc{RigL,
      title={Rigging the Lottery: Making All Tickets Winners}, 
      author={Utku Evci and Trevor Gale and Jacob Menick and Pablo Samuel Castro and Erich Elsen},
      year={2021},
      eprint={1911.11134},
      archivePrefix={arXiv},
      primaryClass={cs.LG},
      url={https://arxiv.org/abs/1911.11134}, 
}

@inproceedings{SRigL,
  author = {Lasby, Mike and Golubeva, Anna and Evci, Utku and Nica, Mihai and Ioannou, Yani},
  booktitle = {{International Conference on Learning Representations (ICLR)}},
  abbr = {ICLR},
  venue = {{Vienna, Austria}},
  eventdate = {2024-05-07/2024-05-11},
  title = {Dynamic Sparse Training with Structured Sparsity},
  year = {2024},
  arxivid = {2305.02299},
  eprint = {2305.02299},
  eprinttype = {arXiv},
  bibtex_show = {true},
  selected = {true}
}

@misc{ModelCompressionSurveyCheng,
      title={A Survey of Model Compression and Acceleration for Deep Neural Networks}, 
      author={Yu Cheng and Duo Wang and Pan Zhou and Tao Zhang},
      year={2020},
      eprint={1710.09282},
      archivePrefix={arXiv},
      primaryClass={cs.LG},
      url={https://arxiv.org/abs/1710.09282}, 
}

@misc{Net2Net,
      title={Net2Net: Accelerating Learning via Knowledge Transfer}, 
      author={Tianqi Chen and Ian Goodfellow and Jonathon Shlens},
      year={2016},
      eprint={1511.05641},
      archivePrefix={arXiv},
      primaryClass={cs.LG},
      url={https://arxiv.org/abs/1511.05641}, 
}

@techreport{CIFAR100Citation,
  title={Learning Multiple Layers of Features from Tiny Images},
  author={Krizhevsky, Alex},
  year={2009},
  institution={University of Toronto},
  type={Technical Report},
  number={TR-2009}
}

@inproceedings{FlowersCitation,
  title={Automated Flower Classification over a Large Number of Classes},
  author={Nilsback, Maria-Elena and Zisserman, Andrew},
  booktitle={Proceedings of the Indian Conference on Computer Vision, Graphics and Image Processing},
  pages={722--729},
  year={2008},
  organization={IEEE}
}

@inproceedings{ImageNet,
  title={ImageNet: A large-scale hierarchical image database},
  author={Deng, Jia and Dong, Wei and Socher, Richard and Li, Li-Jia and Li, Kai and Fei-Fei, Li},
  booktitle={2009 IEEE Conference on Computer Vision and Pattern Recognition},
  pages={248--255},
  year={2009},
  organization={IEEE}
}

@inproceedings{iNaturalistCitation,
  title={The iNaturalist Species Classification and Detection Dataset},
  author={Van Horn, Grant and Mac Aodha, Oisin and Marquis, Trevor and Su, Steve and Haghighi, Mona and Baldridge, Jason and Maji, Subhransu and Perona, Pietro},
  booktitle={Proceedings of the IEEE Conference on Computer Vision and Pattern Recognition (CVPR)},
  pages={8769--8778},
  year={2018},
  organization={IEEE}
}

@misc{PetsDataset,
  title={Cats and Dogs},
  author={Omkar M. Parkhi and Andrea Vedaldi and Andrew Zisserman and C. V. Jawahar},
  howpublished={IEEE Conference on Computer Vision and Pattern Recognition},
  year={2012},
  note={The Oxford-IIIT Pet Dataset}
}

@inproceedings{EvaluatingPruningMethods,
  title={Evaluating pruning methods},
  author={Georg Thimm and Emile Fiesler},
  year={1995},
  booktitle={International Symosium on Artifical Neural Networks},
  url={https://api.semanticscholar.org/CorpusID:11075297}
}

@misc{StructuredMultiHashing,
      title={Structured Multi-Hashing for Model Compression}, 
      author={Elad Eban and Yair Movshovitz-Attias and Hao Wu and Mark Sandler and Andrew Poon and Yerlan Idelbayev and Miguel A. Carreira-Perpinan},
      year={2019},
      eprint={1911.11177},
      archivePrefix={arXiv},
      primaryClass={cs.LG},
      url={https://arxiv.org/abs/1911.11177}, 
}

@misc{TBasis,
      title={T-Basis: a Compact Representation for Neural Networks}, 
      author={Anton Obukhov and Maxim Rakhuba and Stamatios Georgoulis and Menelaos Kanakis and Dengxin Dai and Luc Van Gool},
      year={2021},
      eprint={2007.06631},
      archivePrefix={arXiv},
      primaryClass={cs.LG},
      url={https://arxiv.org/abs/2007.06631}, 
}

@misc{MiniViT,
      title={MiniViT: Compressing Vision Transformers with Weight Multiplexing}, 
      author={Jinnian Zhang and Houwen Peng and Kan Wu and Mengchen Liu and Bin Xiao and Jianlong Fu and Lu Yuan},
      year={2022},
      eprint={2204.07154},
      archivePrefix={arXiv},
      primaryClass={cs.CV},
      url={https://arxiv.org/abs/2204.07154}, 
}

@misc{ALBERT,
      title={ALBERT: A Lite BERT for Self-supervised Learning of Language Representations}, 
      author={Zhenzhong Lan and Mingda Chen and Sebastian Goodman and Kevin Gimpel and Piyush Sharma and Radu Soricut},
      year={2020},
      eprint={1909.11942},
      archivePrefix={arXiv},
      primaryClass={cs.CL},
      url={https://arxiv.org/abs/1909.11942}, 
}

@misc{LessonsParamShareTransformers,
      title={Lessons on Parameter Sharing across Layers in Transformers}, 
      author={Sho Takase and Shun Kiyono},
      year={2023},
      eprint={2104.06022},
      archivePrefix={arXiv},
      primaryClass={cs.CL},
      url={https://arxiv.org/abs/2104.06022}, 
}

@misc{UnderstsandParamSharingTransformers,
      title={Understanding Parameter Sharing in Transformers}, 
      author={Ye Lin and Mingxuan Wang and Zhexi Zhang and Xiaohui Wang and Tong Xiao and Jingbo Zhu},
      year={2023},
      eprint={2306.09380},
      archivePrefix={arXiv},
      primaryClass={cs.LG},
      url={https://arxiv.org/abs/2306.09380}, 
}

@misc{DoRA,
      title={DoRA: Weight-Decomposed Low-Rank Adaptation}, 
      author={Shih-Yang Liu and Chien-Yi Wang and Hongxu Yin and Pavlo Molchanov and Yu-Chiang Frank Wang and Kwang-Ting Cheng and Min-Hung Chen},
      year={2024},
      eprint={2402.09353},
      archivePrefix={arXiv},
      primaryClass={cs.CL},
      url={https://arxiv.org/abs/2402.09353}, 
}

@misc{mishra_accelerating_2021,
	title = {Accelerating {Sparse} {Deep} {Neural} {Networks}},
	url = {http://arxiv.org/abs/2104.08378},
	doi = {10.48550/arXiv.2104.08378},
	urldate = {2023-05-16},
	publisher = {arXiv},
	author = {Mishra, Asit and Latorre, Jorge Albericio and Pool, Jeff and Stosic, Darko and Stosic, Dusan and Venkatesh, Ganesh and Yu, Chong and Micikevicius, Paulius},
	month = apr,
	year = {2021},
	note = {arXiv:2104.08378 [cs]},
}

@misc{neural_magic_neuralmagicdeepsparse_2021,
	title = {DeepSparse Engine: {Sparsity}-aware deep learning inference runtime for {CPUs}},
	url = {https://github.com/neuralmagic/deepsparse},
	urldate = {2024-01-17},
	author = {{Neural Magic}},
	year = {2021},
}

@misc{KaimingHeInit,
      title={Delving Deep into Rectifiers: Surpassing Human-Level Performance on ImageNet Classification}, 
      author={Kaiming He and Xiangyu Zhang and Shaoqing Ren and Jian Sun},
      year={2015},
      eprint={1502.01852},
      archivePrefix={arXiv},
      primaryClass={cs.CV},
      url={https://arxiv.org/abs/1502.01852}, 
}

@misc{AdamW,
      title={Decoupled Weight Decay Regularization}, 
      author={Ilya Loshchilov and Frank Hutter},
      year={2019},
      eprint={1711.05101},
      archivePrefix={arXiv},
      primaryClass={cs.LG},
      url={https://arxiv.org/abs/1711.05101}, 
}

@misc{GradMax,
      title={GradMax: Growing Neural Networks using Gradient Information}, 
      author={Utku Evci and Bart van Merriënboer and Thomas Unterthiner and Max Vladymyrov and Fabian Pedregosa},
      year={2022},
      eprint={2201.05125},
      archivePrefix={arXiv},
      primaryClass={cs.LG},
      url={https://arxiv.org/abs/2201.05125}, 
}

@misc{JaxPruner,
author = {Lee, Joo and Park, Wonpyo and Mitchell, Nicole and Pilault, Jonathan and Obando-Ceron, Johan and Kim, Han-Byul and Lee, Namhoon and Frantar, Elias and Long, Yun and Yazdanbakhsh, Amir and Agrawal, Shivani and Subramanian, Suvinay and Wang, Xin and Kao, Sheng-Chun and Zhang, Xingyao and Gale, Trevor and Bik, Aart and Han, Woohyun and Ferev, Milen and Evci, Utku},
year = {2023},
month = {04},
pages = {},
title = {JaxPruner: A concise library for sparsity research},
doi = {10.48550/arXiv.2304.14082}
}

@misc{SRSTE,
      title={Learning N:M Fine-grained Structured Sparse Neural Networks From Scratch}, 
      author={Aojun Zhou and Yukun Ma and Junnan Zhu and Jianbo Liu and Zhijie Zhang and Kun Yuan and Wenxiu Sun and Hongsheng Li},
      year={2021},
      eprint={2102.04010},
      archivePrefix={arXiv},
      primaryClass={cs.CV},
      url={https://arxiv.org/abs/2102.04010}, 
}

@misc{Gemma2,
      title={Gemma 2: Improving Open Language Models at a Practical Size}, 
      author={Gemma Team and Morgane Riviere and Shreya Pathak and Pier Giuseppe Sessa and Cassidy Hardin and Surya Bhupatiraju and Léonard Hussenot and Thomas Mesnard and Bobak Shahriari and Alexandre Ramé and Johan Ferret and Peter Liu and Pouya Tafti and Abe Friesen and Michelle Casbon and Sabela Ramos and Ravin Kumar and Charline Le Lan and Sammy Jerome and Anton Tsitsulin and Nino Vieillard and Piotr Stanczyk and Sertan Girgin and Nikola Momchev and Matt Hoffman and Shantanu Thakoor and Jean-Bastien Grill and Behnam Neyshabur and Olivier Bachem and Alanna Walton and Aliaksei Severyn and Alicia Parrish and Aliya Ahmad and Allen Hutchison and Alvin Abdagic and Amanda Carl and Amy Shen and Andy Brock and Andy Coenen and Anthony Laforge and Antonia Paterson and Ben Bastian and Bilal Piot and Bo Wu and Brandon Royal and Charlie Chen and Chintu Kumar and Chris Perry and Chris Welty and Christopher A. Choquette-Choo and Danila Sinopalnikov and David Weinberger and Dimple Vijaykumar and Dominika Rogozińska and Dustin Herbison and Elisa Bandy and Emma Wang and Eric Noland and Erica Moreira and Evan Senter and Evgenii Eltyshev and Francesco Visin and Gabriel Rasskin and Gary Wei and Glenn Cameron and Gus Martins and Hadi Hashemi and Hanna Klimczak-Plucińska and Harleen Batra and Harsh Dhand and Ivan Nardini and Jacinda Mein and Jack Zhou and James Svensson and Jeff Stanway and Jetha Chan and Jin Peng Zhou and Joana Carrasqueira and Joana Iljazi and Jocelyn Becker and Joe Fernandez and Joost van Amersfoort and Josh Gordon and Josh Lipschultz and Josh Newlan and Ju-yeong Ji and Kareem Mohamed and Kartikeya Badola and Kat Black and Katie Millican and Keelin McDonell and Kelvin Nguyen and Kiranbir Sodhia and Kish Greene and Lars Lowe Sjoesund and Lauren Usui and Laurent Sifre and Lena Heuermann and Leticia Lago and Lilly McNealus and Livio Baldini Soares and Logan Kilpatrick and Lucas Dixon and Luciano Martins and Machel Reid and Manvinder Singh and Mark Iverson and Martin Görner and Mat Velloso and Mateo Wirth and Matt Davidow and Matt Miller and Matthew Rahtz and Matthew Watson and Meg Risdal and Mehran Kazemi and Michael Moynihan and Ming Zhang and Minsuk Kahng and Minwoo Park and Mofi Rahman and Mohit Khatwani and Natalie Dao and Nenshad Bardoliwalla and Nesh Devanathan and Neta Dumai and Nilay Chauhan and Oscar Wahltinez and Pankil Botarda and Parker Barnes and Paul Barham and Paul Michel and Pengchong Jin and Petko Georgiev and Phil Culliton and Pradeep Kuppala and Ramona Comanescu and Ramona Merhej and Reena Jana and Reza Ardeshir Rokni and Rishabh Agarwal and Ryan Mullins and Samaneh Saadat and Sara Mc Carthy and Sarah Cogan and Sarah Perrin and Sébastien M. R. Arnold and Sebastian Krause and Shengyang Dai and Shruti Garg and Shruti Sheth and Sue Ronstrom and Susan Chan and Timothy Jordan and Ting Yu and Tom Eccles and Tom Hennigan and Tomas Kocisky and Tulsee Doshi and Vihan Jain and Vikas Yadav and Vilobh Meshram and Vishal Dharmadhikari and Warren Barkley and Wei Wei and Wenming Ye and Woohyun Han and Woosuk Kwon and Xiang Xu and Zhe Shen and Zhitao Gong and Zichuan Wei and Victor Cotruta and Phoebe Kirk and Anand Rao and Minh Giang and Ludovic Peran and Tris Warkentin and Eli Collins and Joelle Barral and Zoubin Ghahramani and Raia Hadsell and D. Sculley and Jeanine Banks and Anca Dragan and Slav Petrov and Oriol Vinyals and Jeff Dean and Demis Hassabis and Koray Kavukcuoglu and Clement Farabet and Elena Buchatskaya and Sebastian Borgeaud and Noah Fiedel and Armand Joulin and Kathleen Kenealy and Robert Dadashi and Alek Andreev},
      year={2024},
      eprint={2408.00118},
      archivePrefix={arXiv},
      primaryClass={cs.CL},
      url={https://arxiv.org/abs/2408.00118}, 
}

@misc{Llama3,
      title={The Llama 3 Herd of Models}, 
      author={Aaron Grattafiori and Abhimanyu Dubey and Abhinav Jauhri and Abhinav Pandey and Abhishek Kadian and Ahmad Al-Dahle and Aiesha Letman and Akhil Mathur and Alan Schelten and Alex Vaughan and Amy Yang and Angela Fan and Anirudh Goyal and Anthony Hartshorn and Aobo Yang and Archi Mitra and Archie Sravankumar and Artem Korenev and Arthur Hinsvark and Arun Rao and Aston Zhang and Aurelien Rodriguez and Austen Gregerson and Ava Spataru and Baptiste Roziere and Bethany Biron and Binh Tang and Bobbie Chern and Charlotte Caucheteux and Chaya Nayak and Chloe Bi and Chris Marra and Chris McConnell and Christian Keller and Christophe Touret and Chunyang Wu and Corinne Wong and Cristian Canton Ferrer and Cyrus Nikolaidis and Damien Allonsius and Daniel Song and Danielle Pintz and Danny Livshits and Danny Wyatt and David Esiobu and Dhruv Choudhary and Dhruv Mahajan and Diego Garcia-Olano and Diego Perino and Dieuwke Hupkes and Egor Lakomkin and Ehab AlBadawy and Elina Lobanova and Emily Dinan and Eric Michael Smith and Filip Radenovic and Francisco Guzmán and Frank Zhang and Gabriel Synnaeve and Gabrielle Lee and Georgia Lewis Anderson and Govind Thattai and Graeme Nail and Gregoire Mialon and Guan Pang and Guillem Cucurell and Hailey Nguyen and Hannah Korevaar and Hu Xu and Hugo Touvron and Iliyan Zarov and Imanol Arrieta Ibarra and Isabel Kloumann and Ishan Misra and Ivan Evtimov and Jack Zhang and Jade Copet and Jaewon Lee and Jan Geffert and Jana Vranes and Jason Park and Jay Mahadeokar and Jeet Shah and Jelmer van der Linde and Jennifer Billock and Jenny Hong and Jenya Lee and Jeremy Fu and Jianfeng Chi and Jianyu Huang and Jiawen Liu and Jie Wang and Jiecao Yu and Joanna Bitton and Joe Spisak and Jongsoo Park and Joseph Rocca and Joshua Johnstun and Joshua Saxe and Junteng Jia and Kalyan Vasuden Alwala and Karthik Prasad and Kartikeya Upasani and Kate Plawiak and Ke Li and Kenneth Heafield and Kevin Stone and Khalid El-Arini and Krithika Iyer and Kshitiz Malik and Kuenley Chiu and Kunal Bhalla and Kushal Lakhotia and Lauren Rantala-Yeary and Laurens van der Maaten and Lawrence Chen and Liang Tan and Liz Jenkins and Louis Martin and Lovish Madaan and Lubo Malo and Lukas Blecher and Lukas Landzaat and Luke de Oliveira and Madeline Muzzi and Mahesh Pasupuleti and Mannat Singh and Manohar Paluri and Marcin Kardas and Maria Tsimpoukelli and Mathew Oldham and Mathieu Rita and Maya Pavlova and Melanie Kambadur and Mike Lewis and Min Si and Mitesh Kumar Singh and Mona Hassan and Naman Goyal and Narjes Torabi and Nikolay Bashlykov and Nikolay Bogoychev and Niladri Chatterji and Ning Zhang and Olivier Duchenne and Onur Çelebi and Patrick Alrassy and Pengchuan Zhang and Pengwei Li and Petar Vasic and Peter Weng and Prajjwal Bhargava and Pratik Dubal and Praveen Krishnan and Punit Singh Koura and Puxin Xu and Qing He and Qingxiao Dong and Ragavan Srinivasan and Raj Ganapathy and Ramon Calderer and Ricardo Silveira Cabral and Robert Stojnic and Roberta Raileanu and Rohan Maheswari and Rohit Girdhar and Rohit Patel and Romain Sauvestre and Ronnie Polidoro and Roshan Sumbaly and Ross Taylor and Ruan Silva and Rui Hou and Rui Wang and Saghar Hosseini and Sahana Chennabasappa and Sanjay Singh and Sean Bell and Seohyun Sonia Kim and Sergey Edunov and Shaoliang Nie and Sharan Narang and Sharath Raparthy and Sheng Shen and Shengye Wan and Shruti Bhosale and Shun Zhang and Simon Vandenhende and Soumya Batra and Spencer Whitman and Sten Sootla and Stephane Collot and Suchin Gururangan and Sydney Borodinsky and Tamar Herman and Tara Fowler and Tarek Sheasha and Thomas Georgiou and Thomas Scialom and Tobias Speckbacher and Todor Mihaylov and Tong Xiao and Ujjwal Karn and Vedanuj Goswami and Vibhor Gupta and Vignesh Ramanathan and Viktor Kerkez and Vincent Gonguet and Virginie Do and Vish Vogeti and Vítor Albiero and Vladan Petrovic and Weiwei Chu and Wenhan Xiong and Wenyin Fu and Whitney Meers and Xavier Martinet and Xiaodong Wang and Xiaofang Wang and Xiaoqing Ellen Tan and Xide Xia and Xinfeng Xie and Xuchao Jia and Xuewei Wang and Yaelle Goldschlag and Yashesh Gaur and Yasmine Babaei and Yi Wen and Yiwen Song and Yuchen Zhang and Yue Li and Yuning Mao and Zacharie Delpierre Coudert and Zheng Yan and Zhengxing Chen and Zoe Papakipos and Aaditya Singh and Aayushi Srivastava and Abha Jain and Adam Kelsey and Adam Shajnfeld and Adithya Gangidi and Adolfo Victoria and Ahuva Goldstand and Ajay Menon and Ajay Sharma and Alex Boesenberg and Alexei Baevski and Allie Feinstein and Amanda Kallet and Amit Sangani and Amos Teo and Anam Yunus and Andrei Lupu and Andres Alvarado and Andrew Caples and Andrew Gu and Andrew Ho and Andrew Poulton and Andrew Ryan and Ankit Ramchandani and Annie Dong and Annie Franco and Anuj Goyal and Aparajita Saraf and Arkabandhu Chowdhury and Ashley Gabriel and Ashwin Bharambe and Assaf Eisenman and Azadeh Yazdan and Beau James and Ben Maurer and Benjamin Leonhardi and Bernie Huang and Beth Loyd and Beto De Paola and Bhargavi Paranjape and Bing Liu and Bo Wu and Boyu Ni and Braden Hancock and Bram Wasti and Brandon Spence and Brani Stojkovic and Brian Gamido and Britt Montalvo and Carl Parker and Carly Burton and Catalina Mejia and Ce Liu and Changhan Wang and Changkyu Kim and Chao Zhou and Chester Hu and Ching-Hsiang Chu and Chris Cai and Chris Tindal and Christoph Feichtenhofer and Cynthia Gao and Damon Civin and Dana Beaty and Daniel Kreymer and Daniel Li and David Adkins and David Xu and Davide Testuggine and Delia David and Devi Parikh and Diana Liskovich and Didem Foss and Dingkang Wang and Duc Le and Dustin Holland and Edward Dowling and Eissa Jamil and Elaine Montgomery and Eleonora Presani and Emily Hahn and Emily Wood and Eric-Tuan Le and Erik Brinkman and Esteban Arcaute and Evan Dunbar and Evan Smothers and Fei Sun and Felix Kreuk and Feng Tian and Filippos Kokkinos and Firat Ozgenel and Francesco Caggioni and Frank Kanayet and Frank Seide and Gabriela Medina Florez and Gabriella Schwarz and Gada Badeer and Georgia Swee and Gil Halpern and Grant Herman and Grigory Sizov and Guangyi and Zhang and Guna Lakshminarayanan and Hakan Inan and Hamid Shojanazeri and Han Zou and Hannah Wang and Hanwen Zha and Haroun Habeeb and Harrison Rudolph and Helen Suk and Henry Aspegren and Hunter Goldman and Hongyuan Zhan and Ibrahim Damlaj and Igor Molybog and Igor Tufanov and Ilias Leontiadis and Irina-Elena Veliche and Itai Gat and Jake Weissman and James Geboski and James Kohli and Janice Lam and Japhet Asher and Jean-Baptiste Gaya and Jeff Marcus and Jeff Tang and Jennifer Chan and Jenny Zhen and Jeremy Reizenstein and Jeremy Teboul and Jessica Zhong and Jian Jin and Jingyi Yang and Joe Cummings and Jon Carvill and Jon Shepard and Jonathan McPhie and Jonathan Torres and Josh Ginsburg and Junjie Wang and Kai Wu and Kam Hou U and Karan Saxena and Kartikay Khandelwal and Katayoun Zand and Kathy Matosich and Kaushik Veeraraghavan and Kelly Michelena and Keqian Li and Kiran Jagadeesh and Kun Huang and Kunal Chawla and Kyle Huang and Lailin Chen and Lakshya Garg and Lavender A and Leandro Silva and Lee Bell and Lei Zhang and Liangpeng Guo and Licheng Yu and Liron Moshkovich and Luca Wehrstedt and Madian Khabsa and Manav Avalani and Manish Bhatt and Martynas Mankus and Matan Hasson and Matthew Lennie and Matthias Reso and Maxim Groshev and Maxim Naumov and Maya Lathi and Meghan Keneally and Miao Liu and Michael L. Seltzer and Michal Valko and Michelle Restrepo and Mihir Patel and Mik Vyatskov and Mikayel Samvelyan and Mike Clark and Mike Macey and Mike Wang and Miquel Jubert Hermoso and Mo Metanat and Mohammad Rastegari and Munish Bansal and Nandhini Santhanam and Natascha Parks and Natasha White and Navyata Bawa and Nayan Singhal and Nick Egebo and Nicolas Usunier and Nikhil Mehta and Nikolay Pavlovich Laptev and Ning Dong and Norman Cheng and Oleg Chernoguz and Olivia Hart and Omkar Salpekar and Ozlem Kalinli and Parkin Kent and Parth Parekh and Paul Saab and Pavan Balaji and Pedro Rittner and Philip Bontrager and Pierre Roux and Piotr Dollar and Polina Zvyagina and Prashant Ratanchandani and Pritish Yuvraj and Qian Liang and Rachad Alao and Rachel Rodriguez and Rafi Ayub and Raghotham Murthy and Raghu Nayani and Rahul Mitra and Rangaprabhu Parthasarathy and Raymond Li and Rebekkah Hogan and Robin Battey and Rocky Wang and Russ Howes and Ruty Rinott and Sachin Mehta and Sachin Siby and Sai Jayesh Bondu and Samyak Datta and Sara Chugh and Sara Hunt and Sargun Dhillon and Sasha Sidorov and Satadru Pan and Saurabh Mahajan and Saurabh Verma and Seiji Yamamoto and Sharadh Ramaswamy and Shaun Lindsay and Shaun Lindsay and Sheng Feng and Shenghao Lin and Shengxin Cindy Zha and Shishir Patil and Shiva Shankar and Shuqiang Zhang and Shuqiang Zhang and Sinong Wang and Sneha Agarwal and Soji Sajuyigbe and Soumith Chintala and Stephanie Max and Stephen Chen and Steve Kehoe and Steve Satterfield and Sudarshan Govindaprasad and Sumit Gupta and Summer Deng and Sungmin Cho and Sunny Virk and Suraj Subramanian and Sy Choudhury and Sydney Goldman and Tal Remez and Tamar Glaser and Tamara Best and Thilo Koehler and Thomas Robinson and Tianhe Li and Tianjun Zhang and Tim Matthews and Timothy Chou and Tzook Shaked and Varun Vontimitta and Victoria Ajayi and Victoria Montanez and Vijai Mohan and Vinay Satish Kumar and Vishal Mangla and Vlad Ionescu and Vlad Poenaru and Vlad Tiberiu Mihailescu and Vladimir Ivanov and Wei Li and Wenchen Wang and Wenwen Jiang and Wes Bouaziz and Will Constable and Xiaocheng Tang and Xiaojian Wu and Xiaolan Wang and Xilun Wu and Xinbo Gao and Yaniv Kleinman and Yanjun Chen and Ye Hu and Ye Jia and Ye Qi and Yenda Li and Yilin Zhang and Ying Zhang and Yossi Adi and Youngjin Nam and Yu and Wang and Yu Zhao and Yuchen Hao and Yundi Qian and Yunlu Li and Yuzi He and Zach Rait and Zachary DeVito and Zef Rosnbrick and Zhaoduo Wen and Zhenyu Yang and Zhiwei Zhao and Zhiyu Ma},
      year={2024},
      eprint={2407.21783},
      archivePrefix={arXiv},
      primaryClass={cs.AI},
      url={https://arxiv.org/abs/2407.21783}, 
}

@misc{SlimRedPajama,
  author = {Soboleva, Daria and Al-Khateeb, Faisal and Myers, Robert and Steeves, Jacob R and Hestness, Joel and Dey, Nolan},
  title = {{SlimPajama: A 627B token cleaned and deduplicated version of RedPajama}},
  month = 6,
  year = 2023,
  howpublished = {\url{https://www.cerebras.net/blog/slimpajama-a-627b-token-cleaned-and-deduplicated-version-of-redpajama}},
  url = {https://huggingface.co/datasets/cerebras/SlimPajama-627B},
}

@misc{WikiText2,
      title={Pointer Sentinel Mixture Models}, 
      author={Stephen Merity and Caiming Xiong and James Bradbury and Richard Socher},
      year={2016},
      eprint={1609.07843},
      archivePrefix={arXiv},
      primaryClass={cs.CL},
      url={https://arxiv.org/abs/1609.07843}, 
}

@misc{SVDLLM,
      title={SVD-LLM: Truncation-aware Singular Value Decomposition for Large Language Model Compression},
      author={Xin Wang and Yu Zheng and Zhongwei Wan and Mi Zhang},
      year={2024},
      eprint={2403.07378},
      archivePrefix={arXiv},
      primaryClass={cs.CL},
      url={https://arxiv.org/abs/2403.07378},
}

@misc{SVDLLMv2,
      title={SVD-LLM V2: Optimizing Singular Value Truncation for Large Language Model Compression}, 
      author={Xin Wang and Samiul Alam and Zhongwei Wan and Hui Shen and Mi Zhang},
      year={2025},
      eprint={2503.12340},
      archivePrefix={arXiv},
      primaryClass={cs.CL},
      url={https://arxiv.org/abs/2503.12340}, 
}

@misc{ASVD,
      title={ASVD: Activation-aware Singular Value Decomposition for Compressing Large Language Models},
      author={Zhihang Yuan and Yuzhang Shang and Yue Song and Qiang Wu and Yan Yan and Guangyu Sun},
      year={2023},
      eprint={2312.05821},
      archivePrefix={arXiv},
      primaryClass={cs.CL},
      url={https://arxiv.org/abs/2312.05821},
}

@misc{Llama1,
      title={LLaMA: Open and Efficient Foundation Language Models}, 
      author={Hugo Touvron and Thibaut Lavril and Gautier Izacard and Xavier Martinet and Marie-Anne Lachaux and Timothée Lacroix and Baptiste Rozière and Naman Goyal and Eric Hambro and Faisal Azhar and Aurelien Rodriguez and Armand Joulin and Edouard Grave and Guillaume Lample},
      year={2023},
      eprint={2302.13971},
      archivePrefix={arXiv},
      primaryClass={cs.CL},
      url={https://arxiv.org/abs/2302.13971}, 
}

@software{LMEvalEleuther,
	author = {Lintang Sutawika and Leo Gao and Hailey Schoelkopf and Stella Biderman and Jonathan Tow and Baber Abbasi and ben fattori and Charles Lovering and farzanehnakhaee70 and Jason Phang and Anish Thite and Fazz and Aflah and Niklas Muennighoff and Thomas Wang and sdtblck and nopperl and gakada and tttyuntian and researcher2 and Chris and Julen Etxaniz and Zden{\v e}k Kasner and Khalid and Jeffrey Hsu and AndyZwei and Pawan Sasanka Ammanamanchi and Dirk Groeneveld and Ethan Smith and Eric Tang},
	doi = {10.5281/zenodo.10256836},
	month = dec,
	publisher = {Zenodo},
	title = {EleutherAI/lm-evaluation-harness: Major refactor},
	url = {https://doi.org/10.5281/zenodo.10256836},
	version = {v0.4.0},
	year = 2023}

@misc{hu2021loralowrankadaptationlarge,
      title={LoRA: Low-Rank Adaptation of Large Language Models}, 
      author={Edward J. Hu and Yelong Shen and Phillip Wallis and Zeyuan Allen-Zhu and Yuanzhi Li and Shean Wang and Lu Wang and Weizhu Chen},
      year={2021},
      eprint={2106.09685},
      archivePrefix={arXiv},
      primaryClass={cs.CL},
      url={https://arxiv.org/abs/2106.09685}, 
}

@misc{C4Dataset,
      title={Documenting Large Webtext Corpora: A Case Study on the Colossal Clean Crawled Corpus}, 
      author={Jesse Dodge and Maarten Sap and Ana Marasović and William Agnew and Gabriel Ilharco and Dirk Groeneveld and Margaret Mitchell and Matt Gardner},
      year={2021},
      eprint={2104.08758},
      archivePrefix={arXiv},
      primaryClass={cs.CL},
      url={https://arxiv.org/abs/2104.08758}, 
}

@misc{BasisSharing,
      title={Basis Sharing: Cross-Layer Parameter Sharing for Large Language Model Compression}, 
      author={Jingcun Wang and Yu-Guang Chen and Ing-Chao Lin and Bing Li and Grace Li Zhang},
      year={2024},
      eprint={2410.03765},
      archivePrefix={arXiv},
      primaryClass={cs.CL},
      url={https://arxiv.org/abs/2410.03765}, 
}
